\PassOptionsToPackage{table}{xcolor}
\documentclass[sigconf,screen]{acmart}
\usepackage{multirow}
\usepackage{diagbox}
\usepackage{float}
\usepackage{pifont}
\newcommand{\methodname}{Tex3D}
\definecolor{MethodTint}{RGB}{236,241,248}
\definecolor{DeltaTone}{RGB}{58,128,110}
\definecolor{GainTone}{HTML}{B35656}
\newcommand{\padscore}[1]{\ifdim #1pt<10pt \hphantom{0}#1\else #1\fi}

\newcommand{\gain}[1]{\textcolor{GainTone}{\scriptsize($\uparrow$#1)}}

\newcommand{\scoregain}[2]{\padscore{#1}\,\gain{#2}}

\newcommand{\oursgain}[2]{\cellcolor{MethodTint}\textbf{\padscore{#1}}\,\gain{#2}}
\AtBeginDocument{%
  }

\setcopyright{acmlicensed}
\copyrightyear{2018}
\acmYear{2018}
\acmDOI{XXXXXXX.XXXXXXX}
\acmConference[ACM MM]{Make sure to enter the correct
  conference title from your rights confirmation email}{2026}{Rio de Janeiro, Brazil}
\acmISBN{978-1-4503-XXXX-X/2018/06}

\acmSubmissionID{1311}

\renewcommand\footnotetextcopyrightpermission[1]{} 
\begin{document}

\title{Tex3D: Objects as Attack Surfaces via Adversarial 3D Textures for Vision-Language-Action Models}

\author{%
  Jiawei Chen\textsuperscript{$*$1,2},~
  Simin Huang\textsuperscript{$*$1},~
  Jiawei Du\textsuperscript{3},~
  Shuaihang Chen\textsuperscript{2,5},~
  Yu Tian\textsuperscript{4},~
  Mingjie Wei\textsuperscript{2,5},~
  Chao Yu\textsuperscript{$\dagger$4},~
  Zhaoxia Yin\textsuperscript{$\dagger$1}%
}
\affiliation{%
  \institution{%
    \textsuperscript{$*$}Equal contribution\quad
    \textsuperscript{$\dagger$}Corresponding authors\\[3pt]
    \textsuperscript{1}East China Normal University\quad
    \textsuperscript{2}Zhongguancun Academy\quad
    \textsuperscript{3}CFAR, A*STAR, Singapore\quad
    \textsuperscript{4}Tsinghua University\quad
    \textsuperscript{5}Harbin Institute of Technology%
  }%
  \city{}
  \country{}
}

\begin{abstract}
  Vision-language-action (VLA) models have shown strong performance in robotic manipulation, yet their robustness to physically realizable adversarial attacks remains underexplored. Existing studies reveal vulnerabilities through language perturbations and 2D visual attacks, but these attack surfaces are either less representative of real deployment or limited in physical realism. In contrast, adversarial 3D textures pose a more physically plausible and damaging threat, as they are naturally attached to manipulated objects and are easier to deploy in physical environments. Bringing adversarial 3D textures to VLA systems is nevertheless nontrivial. A central obstacle is that standard 3D simulators do not provide a differentiable optimization path from the VLA objective function back to object appearance, making it difficult to optimize through an end-to-end manner. To address this, we introduce Foreground-Background Decoupling (FBD), which enables differentiable texture optimization through dual-renderer alignment while preserving the original simulation environment. To further ensure that the attack remains effective across long-horizon and diverse viewpoints in the physical world, we propose Trajectory-Aware Adversarial Optimization (TAAO), which prioritizes behaviorally critical frames and stabilizes optimization with a vertex-based parameterization. Built on these designs, we present \textbf{\methodname{}}, the first framework for end-to-end optimization of 3D adversarial textures directly within the VLA simulation environment. Experiments in both simulation and real-robot settings show that Tex3D significantly degrades VLA performance across multiple manipulation tasks, achieving task failure rates of up to 96.7\%. Our empirical results expose critical vulnerabilities of VLA systems to physically grounded 3D adversarial attacks and highlight the need for robustness-aware training. The project page is available at \url{https://vla-attack.github.io/tex3d}.
\end{abstract}



\hypersetup{pdfauthor={Jiawei Chen, Simin Huang, Jiawei Du, Shuaihang Chen, Yu Tian, Mingjie Wei, Chao Yu, Zhaoxia Yin}}
\maketitle

\section{Introduction}
\label{sec:intro}
Vision-language-action (VLA) models~\cite{black2024pi_0,Kim2024OpenVLAAO,brohan2022rt} have achieved remarkable progress in robotic manipulation.
By processing visual observations and natural language instructions in an end-to-end manner, these models directly map multimodal inputs to low-level control signals~\cite{liu2024rdt,zitkovich2023rt}, delivering strong performance across a wide range of complex manipulation tasks.
While VLA models represent a significant step toward general-purpose robotic intelligence, they also introduce potential safety vulnerabilities, particularly against adversarial attacks~\cite{liu2024exploring,wang2025exploring,yang2025mla}.
A malicious actor could manipulate a VLA model into executing incorrect or harmful actions, causing it to perform operations unrelated to the intended task or to exhibit unsafe behaviors.
Assessing the adversarial robustness of VLA models is therefore critical to ensure that deployed systems can respond reliably and safely across varied real-world conditions.

\begin{figure}[t]
\begin{center}
\includegraphics[width=0.95\linewidth]{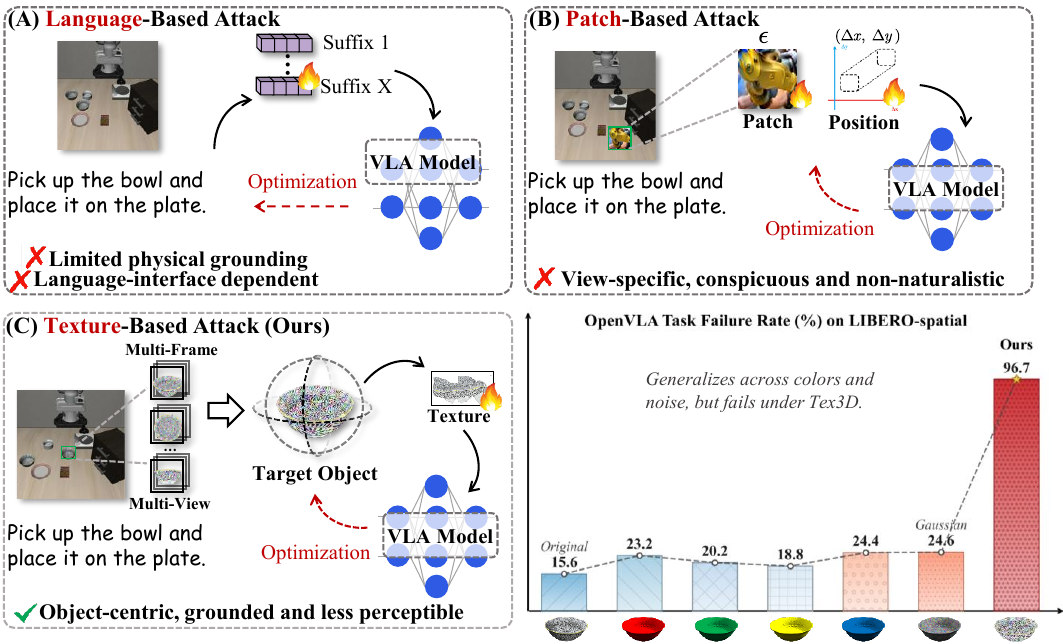}
\end{center}
\caption{Comparison between \methodname{} and existing attack paradigms. Bottom-right: VLA exhibits a certain degree of generalization under color changes and Gaussian noise perturbations, but its task failure rate rises sharply under \methodname{}.} 
\label{fig:first}
\end{figure}
Existing adversarial attacks on VLA models generally fall into two categories.
One line of work~\cite{liu2024exploring,yanSun} targets the language modality by injecting adversarial perturbations into the text instruction, causing the model to output erroneous actions. While effective, these methods are tightly coupled to the language interface.
Another line of work~\cite{wang2025exploring,liu2025eva,luWed,jia2022physical,zhangWed} operates at the 2D visual front-end by affixing adversarial patches onto the input image, inducing task failure through perception-level manipulation. Due to weaker coupling with model-specific interfaces, this line of attack has been more widely studied. However, 2D adversarial patches are inherently view-specific: their effectiveness is contingent on precise viewpoint and pose alignment, which is difficult to guarantee in physical deployment and easy to detect due to their visible, non-naturalistic appearance. Motivated by these limitations, \emph{this paper aims to develop a more physically grounded and less perceptible attack surface: adversarial 3D textures}. Unlike 2D patches, adversarial 3D textures are directly bound to object surfaces, making them inherently robust to viewpoint and object pose variations during embodied interaction, and more naturally integrated into the object's appearance. This object-centric attack surface is particularly suitable for embodied manipulation and real-world deployment.

\begin{figure*}[t]
\begin{center}
\includegraphics[width=0.9\linewidth]{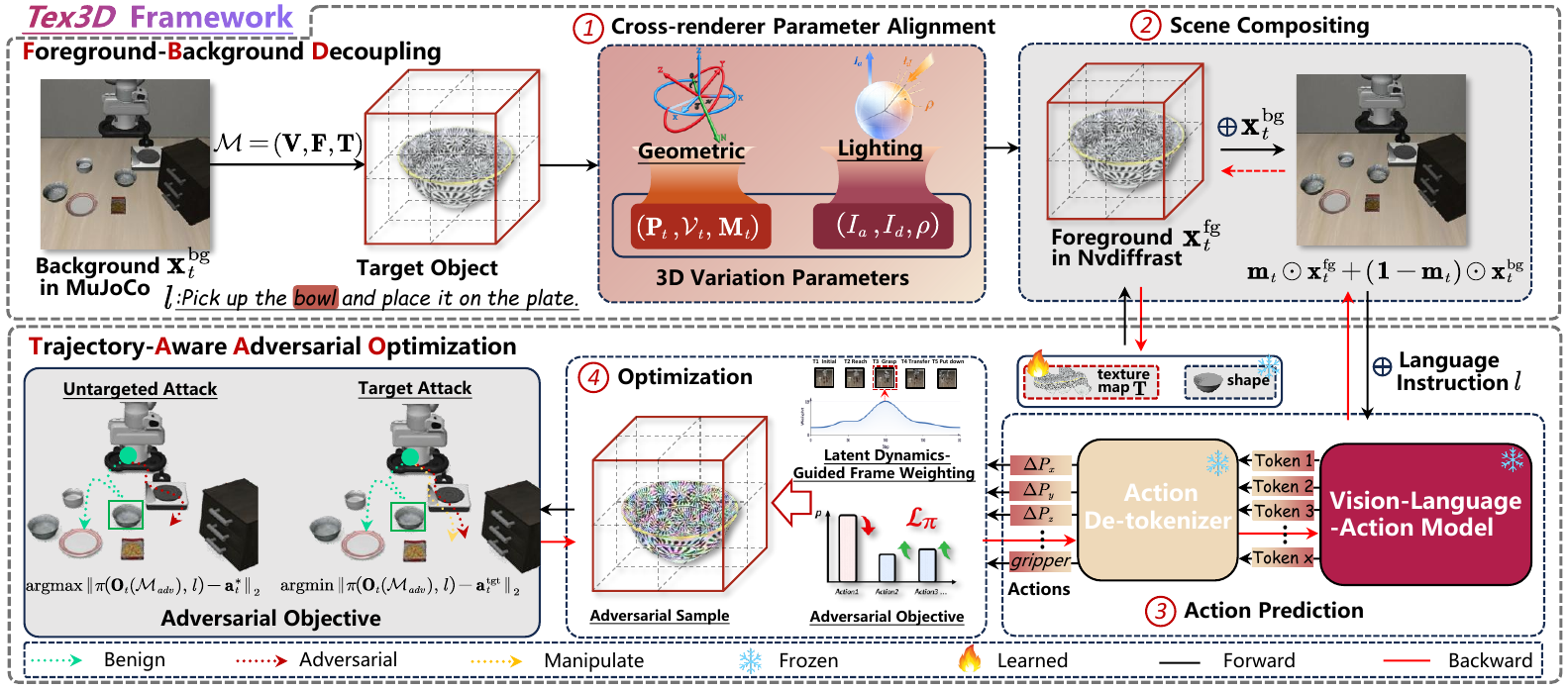}
\end{center}
\caption{Overview of \methodname{}. FBD renders the background in MuJoCo and the target object in Nvdiffrast, with cross-renderer alignment of geometric parameters $(\mathbf{P}_t,\mathcal{V}_t,\mathbf{M}_t)$ and lighting parameters $(I_a,I_d,\rho)$ for photometrically consistent scene composition. The composited observation is fed into the frozen VLA model, and gradients from untargeted or targeted objectives are back-propagated to directly optimize the object texture. TAAO further applies dynamics-guided weighting over critical frames, enabling temporally effective adversarial 3D texture optimization over complex long-horizon manipulation trajectories.} 
\label{fig:frame}
\end{figure*}
However, achieving this goal requires overcoming two sequential challenges. The first is the \emph{lack of differentiability with respect to object appearance}. Common embodied simulation platforms such as MuJoCo~\cite{todorov2012mujoco} (the physics engine underlying LIBERO~\cite{liu2023libero}) are non-differentiable w.r.t. 3D object appearance. Since they expose VLA models only through action-level interactions, gradients cannot be propagated back to object textures, making direct optimization of 3D physical perturbations intractable. As a result, there is currently no straightforward way to optimize adversarial 3D textures in such environments. Once such a differentiable optimization pathway is established, a second challenge arises: \emph{maintaining adversarial effectiveness over long temporal sequences}. Existing 3D texture attacks are mainly developed for non-embodied settings~\cite{xiao2019meshadv,huang2024towards,suryanto2023active} and do not account for the long sequential nature of VLA inference. Even simple tasks span hundreds of frames, and not all frames contribute equally to the model's decisions. A perturbation effective at one timestep may have diminished impact at another, making it hard to sustain adversarial influence consistently over the full trajectory.

To address this, we propose \textbf{\methodname{}}, an adversarial texture attack framework for VLA models. We first introduce \textbf{Foreground-Background Decoupling} (FBD), which establishes a differentiable optimization pathway through cross-renderer parameter alignment and scene compositing. Specifically, MuJoCo renders the background while the target object is rendered in Nvdiffrast~\cite{laine2020modular}; the aligned MVP transforms and lighting parameters ($I_a$, $I_d$, $\rho$) ensure both spatial and photometric coherence across the two renderers. This enables gradients to flow back to the texture map without reconstructing the full simulation stack. Building on this differentiable pipeline, we further propose \textbf{Trajectory-Aware Adversarial Optimization} (TAAO) which consists of two coupled components: (i) identifying behaviorally critical frames via the latent velocity and acceleration of the observation sequence (computed using central differences on a pre-trained visual encoder output) and weighting these frames accordingly via a temperature-scaled softmax; and (ii) jointly reparameterizing the adversarial texture as per-vertex color attributes to constrain optimization to a smooth, low-rank manifold, reducing overfitting and improving attack effectiveness. Building on TAAO, we further instantiate two attack strategies, including untargeted and targeted attacks, to effectively simulate a broad range of adversarial scenarios. Together, these components enable fully end-to-end optimization of physically grounded 3D adversarial textures directly within the embodied simulation loop.

Our main contributions are summarized as follows:
\ding{182}~To the best of our knowledge, \methodname{} is the first framework enabling end to end optimization of 3D adversarial textures directly within a VLA simulation environment.
\ding{183}~We introduce two key techniques, Foreground Background Decoupling (FBD) and Trajectory Aware Adversarial Optimization (TAAO), which enable differentiable and temporally consistent optimization of adversarial 3D textures.
\ding{184}~We conduct extensive evaluations in simulation and real robot settings across four manipulation task suites, showing \methodname{} achieves task failure rates of up to 96.7\%, revealing critical vulnerabilities of current VLA models to physically grounded 3D adversarial attacks.

\section{Related Work}
\subsection{Vision-Language-Action Models}
Recent advances in large vision-language models (LVLMs) have driven the development of VLA models, which unify perception, language grounding, and action generation for robotic control~\cite{brohan2022rt,zitkovich2023rt,team2024octo,Kim2024OpenVLAAO}. Existing methods can be broadly categorized into autoregressive, diffusion-based, and hybrid paradigms~\cite{brohan2022rt,zitkovich2023rt,team2024octo,Kim2024OpenVLAAO,liu2024rdt,black2024pi_0,wen2025tinyvla,shukor2025smolvla,liu2025hybridvla}. Among them, OpenVLA~\cite{Kim2024OpenVLAAO,o2024open,karamcheti2024prismatic} and $\pi_0$~\cite{black2024pi_0} are two representative open-source VLA models, following token-based autoregressive prediction and flow-matching diffusion policies, respectively. Recent extensions such as OpenVLA-OFT~\cite{liang2025finetuningvisionlanguageactionmodels}, $\pi_{0.5}$~\cite{intelligence2025pi_}, and other enhanced VLA frameworks~\cite{zheng2024tracevla,qu2025spatialvlaexploringspatial,li2024cogact} further improve efficiency, spatial-temporal reasoning, and generalization. Despite strong performance, these models remain highly vulnerable to adversarial attacks~\cite{liu2024exploring,dupuy2025embodiedredteaming,yanSun,jones2025adversarial}, as even small visual perturbations can propagate through the tightly coupled perception-language-action pipeline and trigger potentially unsafe or erroneous actions.
\subsection{Adversarial Attacks in VLA Models}
Adversarial attacks~\cite{kurakin2016adversarial,shafahi2019adversarial,chen2023advfas,chen2025autobreach} are widely used to probe vulnerabilities in VLA models, where small perturbations can lead to task failures or unsafe behaviors~\cite{liu2024exploring,dupuy2025embodiedredteaming,wang2025robosafe}. Existing VLA attacks can generally be divided into two categories: (1) language-based attacks, which manipulate instructions to subtly influence action generation~\cite{liu2024exploring,dupuy2025embodiedredteaming,yanSun}, and (2) patch-based attacks, which apply adversarial perturbations to input images, typically via 2D patches that significantly degrade model performance and can often transfer across settings~\cite{wang2025exploring,liu2025eva,luWed,jia2022physical,zhangWed}. Compared to language-based methods, visual attacks are more practical due to their direct impact on perception and physical realizability~\cite{fei2025libero}. However, existing 2D patch attacks are inherently viewpoint-dependent, requiring precise alignment between camera pose and object geometry, and their conspicuous appearance makes them easier to detect. While prior work has explored 3D adversarial perturbations using point clouds, meshes, and neural rendering to improve cross-view robustness~\cite{xiao2019meshadv,xiang2019pointcloud,li2023adv3d}, these approaches are not directly applicable to VLA systems.
To effectively address these limitations, we explore adversarial 3D textures bound to target object surfaces, enabling more robust, physically realizable, and less perceptible attacks.
\section{Methodology}
\subsection{Problem Formulation}
\label{sec:formula}
We consider embodied simulation environments such as MuJoCo, where the visual scene consists of $K$ 3D objects. Each object is modeled as a textured mesh $\mathcal{M} = (\mathbf{V}, \mathbf{F}, \mathbf{T})$, with vertex coordinates $\mathbf{V}$, triangle face indices $\mathbf{F}$, and a texture map $\mathbf{T} \in \mathbb{R}^{H_t \times W_t \times 3}$ that encodes its surface appearance. The scene is rendered at each timestep $t$ into an RGB observation $\mathbf{O}_t \in \mathbb{R}^{H \times W \times 3}$, which serves as the visual input to the VLA model $\pi$. The model takes $\mathbf{O}_t$ and a language instruction $l$ and outputs actions $\mathbf{a}_t \in \mathbb{R}^{d}$, i.e., $\pi(\mathbf{O}_t, l) = \mathbf{a}_t$.

\noindent\textbf{Optimization Objective.}
The goal of \methodname{} is to find an adversarial texture map $\mathbf{T}_{adv}$ that encodes a texture capable of disrupting the VLA model's action outputs, while leaving the object's geometry and semantic identity intact.
Since VLA inference is inherently sequential, even a simple task unfolds over hundreds of timesteps; the adversarial texture must therefore sustain its effect across the full trajectory rather than a single observation.
Moreover, to improve robustness against viewpoint changes, we optimize the texture jointly over $M$ sampled views at each timestep. Let $\mathbf{O}_{t,m}(\mathcal{M})$ denote the observation of mesh $\mathcal{M}$ at timestep $t$ under the $m$-th sampled view, where $m \in \{1,\dots,M\}$.
Let $\mathbf{a}^*_{t,m} = \pi(\mathbf{O}_{t,m}(\mathcal{M}), l)$ be the corresponding reference action under the clean mesh.
\methodname{} thus seeks $\mathbf{T}_{adv}$ that maximizes the action deviation in expectation jointly over both the task trajectory $\mathcal{T}$ and sampled views during optimization:
\begin{equation}
  \begin{split}
  \mathbf{T}_{adv} = \underset{\mathbf{T}_{adv}}{\arg\max}
  \;\mathop{\mathbb{E}}\limits_{t \sim \mathcal{T},\, m \sim \mathcal{U}}\!
  \left[
    \mathcal{L}_{\pi}\!\left(
      \pi\!\left(\mathbf{O}_{t,m}(\mathcal{M}_{adv}),\, l\right),\;
      \mathbf{a}^*_{t,m}
    \right)
    \right], \\
    \text{where } \mathcal{M}_{adv} = \left(\mathbf{V},\, \mathbf{F},\, \mathbf{T}_{adv}\right).
  \end{split}
  \label{eq:objective}
\end{equation}

To enable end-to-end optimization of $\mathbf{T}_{adv}$ directly within the simulator, we design FBD, a differentiable rendering pipeline that decouples the adversarial object from its static background, allowing gradients to directly flow back to the texture map $\mathbf{T}$; details are given in Sec.~\ref{method:render}.
To realize the trajectory-level expectation $\mathbb{E}_{t \sim \mathcal{T}}$ in Eq.~\eqref{eq:objective} both effectively and efficiently, we further propose TAAO, which automatically identifies behaviorally critical frames via latent dynamics of the observation sequence and concentrates optimization on them; details are given in Sec.~\ref{method:optimization}.
Finally, to improve the robustness of our framework under real-world conditions, we incorporate an EoT scheme to bridge the gap between digital-domain optimization and physical deployment, as detailed in Sec.~\ref{method:physical_attack}.

\subsection{Differentiable Rendering via Foreground--Background Decoupling}
\label{method:render}

Common embodied simulators such as MuJoCo typically serve primarily as physics engines and do not expose a gradient path through 3D object appearance.
We propose \emph{Foreground-Background Decoupling} (FBD): MuJoCo is kept running unmodified, while the target object is additionally rendered in the differentiable renderer Nvdiffrast fully in parallel.
At each step, the relevant rendering parameters (object pose, camera viewpoint, and lighting) are relayed from MuJoCo to Nvdiffrast for precisely synchronized rendering; corresponding gradients are then computed through Nvdiffrast to update the adversarial texture, which is subsequently applied back to the object in MuJoCo for the next simulation step.
This design combines the strengths of both renderers: Nvdiffrast provides differentiable gradients for texture optimization, while MuJoCo retains full control over the simulation environment and the VLA model interaction interface, which Nvdiffrast alone cannot provide.

\noindent\textbf{\ding{182}~Environmental background rendering.}
At timestep $t$, MuJoCo renders the complete scene under physically accurate simulation, capturing the full environmental context (robot, tabletop, and surrounding objects) as the faithful background reference:
\begin{equation}
  \mathbf{x}_t^{\mathrm{bg}} \in [0,1]^{3 \times H \times W}.
  \label{eq:bg}
\end{equation}

\noindent\textbf{\ding{183}~Target foreground rendering.}
We adopt Nvdiffrast, a high-performance differentiable renderer developed by NVIDIA that provides GPU-accelerated primitive operations based on rasterization, enabling efficient gradient computation over 3D mesh attributes including texture.
The target object mesh $\mathcal{M}_{adv} = (\mathbf{V}, \mathbf{F}, \mathbf{T}_{adv})$ is rendered under rendering parameters aligned with MuJoCo (see cross-renderer geometric alignment below).
This produces the foreground image $\mathbf{x}_t^{\mathrm{fg}}(\mathbf{T}_{adv})$ with gradients available with respect to $\mathbf{T}_{adv}$, and a silhouette mask $\mathbf{m}_t \in \{0,1\}^{H \times W}$ pre-computed from MuJoCo's native renderer that identifies which pixels correspond to the target object and which belong to the background.

\noindent\textbf{\ding{184}~Scene compositing.}
After foreground object rendering, the adversarial observation is explicitly assembled by substituting the object with the adversarial texture seamlessly back into the scene:
\begin{equation}
  \mathbf{O}_t(\mathcal{M}_{adv}) = \mathbf{m}_t \odot \mathbf{x}_t^{\mathrm{fg}}(\mathbf{T}_{adv})
  + (\mathbf{1} - \mathbf{m}_t) \odot \mathbf{x}_t^{\mathrm{bg}},
  \label{eq:compositing}
\end{equation}
where $\odot$ denotes element-wise multiplication.
This is equivalent to applying $\mathbf{T}_{adv}$ to the object in MuJoCo, while maintaining a differentiable path from $\mathbf{T}_{adv}$ back through the action loss.
$\mathbf{O}_t(\mathcal{M}_{adv})$ is fed to the VLA model, which produces the action $\mathbf{a}_t$, completing the gradient chain $\mathbf{T}_{adv} \!\to\! \mathbf{x}_t^{\mathrm{fg}} \!\to\! \mathbf{O}_t \!\to\! \pi \!\to\! \mathcal{L}_\pi$ required by Eq.~\eqref{eq:objective}.

\noindent\textbf{Cross-renderer parameter alignment.}
To ensure that Nvdiffrast renders the target object consistently with its appearance in MuJoCo, we synchronize the rendering parameters of both renderers at each timestep.
\underline{\emph{Geometric alignment.}}
Let $\mathbf{M}_t \in \mathbb{R}^{4 \times 4}$ be the model matrix of the target object, encoding its 3D pose in world coordinates.
The view matrix $\mathcal{V}_t$ and projection matrix $\mathbf{P}_t$ are derived from the simulator's camera pose and intrinsic parameters.
Following the widely used standard MVP transform in computer graphics, the clip-space coordinate of each vertex is obtained as:
\begin{equation}
  \mathbf{v}_{i,t}^{\mathrm{clip}} = \mathbf{C}_t\,\tilde{\mathbf{v}}_i, \quad
  \mathbf{C}_t = \mathbf{P}_t\, \mathcal{V}_t\, \mathbf{M}_t \in \mathbb{R}^{4 \times 4},
  \label{eq:clip_transform}
\end{equation}
where $\tilde{\mathbf{v}}_i \in \mathbb{R}^4$ is the precisely defined homogeneous form of the $i$-th vertex $\mathbf{v}_i \in \mathbf{V}$; $\mathbf{M}_t$ transforms the vertex from object space to world space; $\mathcal{V}_t$ further projects it into camera space; and $\mathbf{P}_t$ maps it to clip space.
Nvdiffrast subsequently takes $\{\mathbf{v}_{i,t}^{\mathrm{clip}}\}$ as input for rasterization and texture interpolation.
Through this transform, the spatial placement of the target object in Nvdiffrast is precisely aligned with its counterpart in MuJoCo, ensuring geometric consistency across the two renderers.
\underline{\emph{Lighting alignment.}}
Beyond geometry, we explicitly align the lighting conditions of the two renderers.
The ambient light intensity $I_a$, diffuse light intensity $I_d$, and material reflectance $\rho$ of the target object are read from MuJoCo's scene configuration and applied to the Nvdiffrast shading pipeline.
Through this joint alignment, the viewpoint, pose, and surface shading of the target object are kept faithfully consistent between the two renderers, yielding a photometrically coherent composited observation.

\begin{table*}[t]
  \caption{Task failure rates (\%) on four LIBERO task variants under untargeted and targeted settings. ``No Attack'' denotes clean evaluation, ``Gaussian'': random Gaussian noise, ``Single-frame'': one-frame perturbation, ``Vertex Param.'': our parameterized texture attack without temporal consistency, ``Tex+Temp.'': the texture-only temporal variant, and \methodname{}: our full method.}
  \label{tab:main}
  \centering
  \footnotesize
  \setlength{\tabcolsep}{3pt}
  \resizebox{0.95\textwidth}{!}{%
  \begin{tabular}{ll cccccc cccccc}
    \toprule
    \multirow{2}{*}{Model} & \multirow{2}{*}{Task}
      & \multicolumn{6}{c}{Untargeted Attack}
      & \multicolumn{6}{c}{Targeted Attack} \\
    \cmidrule(lr){3-8}\cmidrule(lr){9-14}
    & & No Attack & Gaussian & Single-frame & Vertex Param. & Tex+Temp. & \cellcolor{MethodTint}\textbf{\methodname{}}
      & No Attack & Gaussian & Single-frame & Vertex Param. & Tex+Temp. & \cellcolor{MethodTint}\textbf{\methodname{}} \\
    \midrule
    \multirow{5}{*}{OpenVLA~\cite{Kim2024OpenVLAAO}}
      & Spatial & 15.6 & \scoregain{24.6}{9.0} & \scoregain{75.1}{59.5} & \scoregain{80.5}{64.9} & \scoregain{90.3}{74.7} & \oursgain{95.8}{80.2}
      & 15.6 & \scoregain{24.6}{9.0} & \scoregain{82.4}{66.8} & \scoregain{86.5}{70.9} & \scoregain{93.2}{77.6} & \oursgain{96.7}{81.1} \\
      & Object  & 11.8 & \scoregain{18.8}{7.0} & \scoregain{62.1}{50.3} & \scoregain{69.8}{58.0} & \scoregain{74.6}{62.8} & \oursgain{81.1}{69.3}
      & 11.8 & \scoregain{18.8}{7.0} & \scoregain{70.9}{59.1} & \scoregain{74.8}{63.0} & \scoregain{81.1}{69.3} & \oursgain{85.5}{73.7} \\
      & Goal    & 23.6 & \scoregain{28.8}{5.2} & \scoregain{65.5}{41.9} & \scoregain{70.7}{47.1} & \scoregain{79.1}{55.5} & \oursgain{84.8}{61.2}
      & 23.6 & \scoregain{28.8}{5.2} & \scoregain{71.5}{47.9} & \scoregain{74.2}{50.6} & \scoregain{81.6}{58.0} & \oursgain{86.9}{63.3} \\
      & Long    & 45.4 & \scoregain{52.2}{6.8} & \scoregain{73.2}{27.8} & \scoregain{80.9}{35.5} & \scoregain{87.8}{42.4} & \oursgain{90.9}{45.5}
      & 45.4 & \scoregain{52.2}{6.8} & \scoregain{79.6}{34.2} & \scoregain{84.3}{38.9} & \scoregain{90.5}{45.1} & \oursgain{92.8}{47.4} \\
      & Avg.    & 24.1 & \scoregain{31.1}{7.0} & \scoregain{69.0}{44.9} & \scoregain{75.5}{51.4} & \scoregain{82.9}{58.8} & \oursgain{88.1}{64.0}
      & 24.1 & \scoregain{31.1}{7.0} & \scoregain{76.1}{52.0} & \scoregain{79.9}{55.8} & \scoregain{86.6}{62.5} & \oursgain{90.5}{66.4} \\
    \midrule
    \multirow{5}{*}{OpenVLA-OFT~\cite{liang2025finetuningvisionlanguageactionmodels}}
      & Spatial &  3.8 & \scoregain{ 8.8}{5.0} & \scoregain{69.9}{66.1} & \scoregain{72.2}{68.4} & \scoregain{74.9}{71.1} & \oursgain{78.6}{74.8}
      &  3.8 & \scoregain{ 8.8}{5.0} & \scoregain{70.8}{67.0} & \scoregain{74.4}{70.6} & \scoregain{76.9}{73.1} & \oursgain{80.9}{77.1} \\
      & Object  &  1.7 & \scoregain{ 3.6}{1.9} & \scoregain{57.4}{55.7} & \scoregain{64.9}{63.2} & \scoregain{67.7}{66.0} & \oursgain{70.6}{68.9}
      &  1.7 & \scoregain{ 3.6}{1.9} & \scoregain{63.7}{62.0} & \scoregain{69.2}{67.5} & \scoregain{71.2}{69.5} & \oursgain{76.4}{74.7} \\
      & Goal    &  3.8 & \scoregain{ 5.4}{1.6} & \scoregain{61.3}{57.5} & \scoregain{68.1}{64.3} & \scoregain{72.4}{68.6} & \oursgain{76.8}{73.0}
      &  3.8 & \scoregain{ 5.4}{1.6} & \scoregain{64.6}{60.8} & \scoregain{70.9}{67.1} & \scoregain{73.9}{70.1} & \oursgain{79.8}{76.0} \\
      & Long    &  9.3 & \scoregain{ 10.9}{1.6} & \scoregain{65.5}{56.2} & \scoregain{68.0}{58.7} & \scoregain{73.6}{64.3} & \oursgain{78.2}{68.9}
      &  9.3 &\scoregain{ 10.9}{1.6} & \scoregain{68.9}{59.6} & \scoregain{71.7}{62.4} & \scoregain{75.8}{66.5} & \oursgain{80.2}{70.9} \\
      & Avg.    &  4.7 & \scoregain{ 6.5}{1.8} & \scoregain{63.5}{58.8} & \scoregain{68.3}{63.6} & \scoregain{72.1}{67.4} & \oursgain{76.0}{71.3}
      &  4.7 & \scoregain{ 6.5}{1.8} & \scoregain{67.0}{62.3} & \scoregain{71.5}{66.8} & \scoregain{74.4}{69.7} & \oursgain{79.3}{74.6} \\
    \midrule
\multirow{5}{*}{$\pi_0$~\cite{black2024pi_0}}
  & Spatial &  3.5 & \scoregain{11.8}{8.3} & \scoregain{54.7}{51.2} & \scoregain{59.4}{55.9} & \scoregain{65.2}{61.7} & \oursgain{74.9}{71.4}
  &  3.5 & \scoregain{11.8}{8.3} & \scoregain{58.8}{55.3} & \scoregain{63.2}{59.7} & \scoregain{69.5}{66.0} & \oursgain{75.9}{72.4} \\
  & Object  &  2.3 & \scoregain{8.6}{6.3} & \scoregain{48.6}{46.3} & \scoregain{55.4}{53.1} & \scoregain{59.3}{57.0} & \oursgain{68.9}{66.6}
  &  2.3 & \scoregain{8.6}{6.3} & \scoregain{52.5}{50.2} & \scoregain{57.2}{54.9} & \scoregain{62.6}{60.3} & \oursgain{70.2}{67.9} \\
  & Goal    &  5.2 & \scoregain{9.6}{4.4} & \scoregain{53.0}{47.8} & \scoregain{58.3}{53.1} & \scoregain{63.1}{57.9} & \oursgain{70.3}{65.1}
  &  5.2 & \scoregain{9.6}{4.4} & \scoregain{56.2}{51.0} & \scoregain{60.5}{55.3} & \scoregain{68.6}{63.4} & \oursgain{73.4}{68.2} \\
  & Long    &  7.2 & \scoregain{12.7}{5.5} & \scoregain{54.2}{47.0} & \scoregain{57.7}{50.5} & \scoregain{64.5}{57.3} & \oursgain{72.9}{65.7}
  &  7.2 & \scoregain{12.7}{5.5} & \scoregain{57.2}{50.0} & \scoregain{61.4}{54.2} & \scoregain{68.4}{61.2} & \oursgain{73.7}{66.5} \\
  & Avg.    &  4.6 & \scoregain{10.7}{6.1} & \scoregain{52.6}{48.0} & \scoregain{57.7}{53.1} & \scoregain{63.0}{58.4} & \oursgain{71.8}{67.2}
  &  4.6 & \scoregain{10.7}{6.1} & \scoregain{56.2}{51.6} & \scoregain{60.6}{56.0} & \scoregain{67.3}{62.7} & \oursgain{73.3}{68.7} \\
    \midrule
    \multirow{5}{*}{$\pi_{0.5}$~\cite{intelligence2025pi_}}
  & Spatial &  1.2 & \scoregain{7.7}{6.5} & \scoregain{47.1}{45.9} & \scoregain{52.5}{51.3} & \scoregain{60.1}{58.9} & \oursgain{71.8}{70.6}
  &  1.2 & \scoregain{7.7}{6.5} & \scoregain{49.3}{48.1} & \scoregain{55.2}{54.0} & \scoregain{63.2}{62.0} & \oursgain{72.9}{71.7} \\
  & Object  &  1.8 & \scoregain{6.9}{5.1} & \scoregain{39.7}{37.9} & \scoregain{46.0}{44.2} & \scoregain{52.4}{50.6} & \oursgain{65.2}{63.4}
  &  1.8 & \scoregain{6.9}{5.1} & \scoregain{41.1}{39.3} & \scoregain{48.2}{46.4} & \scoregain{54.3}{52.5} & \oursgain{68.3}{66.5} \\
  & Goal    &  2.0 & \scoregain{5.5}{3.5} & \scoregain{44.7}{42.7} & \scoregain{50.9}{48.9} & \scoregain{57.7}{55.7} & \oursgain{69.7}{67.7}
  &  2.0 & \scoregain{5.5}{3.5} & \scoregain{47.5}{45.5} & \scoregain{53.2}{51.2} & \scoregain{60.8}{58.8} & \oursgain{71.3}{69.3} \\
  & Long    &  6.0 & \scoregain{9.3}{3.3} & \scoregain{50.0}{44.0} & \scoregain{55.2}{49.2} & \scoregain{63.0}{57.0} & \oursgain{70.6}{64.6}
  &  6.0 & \scoregain{9.3}{3.3} & \scoregain{51.8}{45.8} & \scoregain{56.8}{50.8} & \scoregain{65.1}{59.1} & \oursgain{72.1}{66.1} \\
  & Avg.    &  2.8 & \scoregain{7.4}{4.6} & \scoregain{45.4}{42.6} & \scoregain{51.2}{48.4} & \scoregain{58.3}{55.5} & \oursgain{69.3}{66.5}
  &  2.8 & \scoregain{7.4}{4.6} & \scoregain{47.4}{44.6} & \scoregain{53.4}{50.6} & \scoregain{60.9}{58.1} & \oursgain{71.2}{68.4} \\
    \bottomrule
  \end{tabular}%
  }
\end{table*}

\begin{table}[t]
  \centering
  \caption{Task failure rates (\%). \textit{Top}: dual-renderer fidelity of OpenVLA; values in parentheses show increases from MuJoCo direct rendering. \textit{Bottom}: cross-model transferability; increases are relative to the target model's clean performance.}
  \label{tab:analysis}
  \footnotesize
  \setlength{\tabcolsep}{4pt}
  \resizebox{0.95\linewidth}{!}{
  \begin{tabular}{lcccc}
    \toprule
    & Spatial & Object & Goal & Long \\
    \midrule
    \multicolumn{5}{l}{\textit{Dual-Renderer Fidelity}} \\
    \midrule
    MuJoCo (direct)      & 15.6 & 11.8 & 23.6 & 45.4 \\
    MuJoCo + nvdiffrast  & \scoregain{16.8}{1.2} & \scoregain{12.2}{0.4} & \scoregain{24.4}{0.8} & \scoregain{45.8}{0.4} \\
    \midrule
    \multicolumn{5}{l}{\textit{Cross-Model Transferability}} \\
    \midrule
    OpenVLA $\to$ OpenVLA-OFT & \scoregain{70.6}{66.8} & \scoregain{61.5}{59.8} & \scoregain{65.4}{61.6} & \scoregain{69.1}{59.8} \\
    OpenVLA-OFT $\to$ OpenVLA & \scoregain{75.3}{59.7} & \scoregain{70.2}{58.4} & \scoregain{73.4}{49.8} & \scoregain{75.7}{30.3} \\
    $\pi_0$ $\to$ $\pi_{0.5}$ & \scoregain{58.4}{57.2} & \scoregain{49.2}{47.4} & \scoregain{54.1}{52.1} & \scoregain{57.6}{51.6} \\
    $\pi_{0.5}$ $\to$ $\pi_0$ & \scoregain{63.7}{60.2} & \scoregain{55.8}{53.5} & \scoregain{60.5}{55.3} & \scoregain{62.3}{55.1} \\
    OpenVLA $\to$ $\pi_0$ & \scoregain{41.2}{37.7} & \scoregain{34.5}{32.2} & \scoregain{38.9}{33.7} & \scoregain{44.1}{36.9} \\
    OpenVLA $\to$ $\pi_{0.5}$ & \scoregain{32.6}{31.4} & \scoregain{27.8}{26.0} & \scoregain{31.4}{29.4} & \scoregain{36.5}{30.5} \\
    $\pi_0$ $\to$ OpenVLA & \scoregain{51.2}{35.6} & \scoregain{44.8}{33.0} & \scoregain{52.4}{28.8} & \scoregain{61.5}{16.1} \\
    $\pi_0$ $\to$ OpenVLA-OFT & \scoregain{43.6}{39.8} & \scoregain{37.2}{35.5} & \scoregain{41.5}{37.7} & \scoregain{48.4}{39.1} \\
    \bottomrule
  \end{tabular}}
\end{table}

\subsection{Trajectory-Aware Adversarial Optimization}
\label{method:optimization}

As clearly established in Sec.~\ref{sec:formula}, effective adversarial textures must consistently exert influence across the full manipulation trajectory rather than at isolated frames.
We address this by proposing trajectory-aware adversarial optimization (TAAO), which exploits task structure to focus adversarial pressure where it matters most.

\noindent\textbf{Latent dynamics-guided frame weighting.}
Frames at which the robot undergoes behavioral transitions (e.g., the onset of grasping or lift-off) are disproportionately influential on task success.
We identify such frames through the latent dynamics of the observation sequence, without relying on any simulator-specific state.

\noindent\emph{Latent feature extraction.}
At each timestep $t$, we first extract a compact latent representation via a pre-trained visual encoder~\cite{esser2021taming} $E$:
\begin{equation}
  \mathbf{f}_t = E(\mathbf{O}_t) \in \mathbb{R}^{d_f}.
\end{equation}

\noindent\emph{Latent velocity and acceleration.}
We estimate the temporal rate and rate-of-change of feature variation using central differences:
\begin{equation}
  v_t = \|\mathbf{f}_{t+1} - \mathbf{f}_{t-1}\|_2 / 2, \qquad \alpha_t = |v_t - v_{t-1}|,
  \label{eq:latent_dynamics}
\end{equation}
and jointly normalize both quantities over the trajectory:
\begin{equation}
  \hat{v}_t = \frac{v_t - v_{\min}}{v_{\max} - v_{\min}}, \qquad
  \hat{\alpha}_t = \frac{\alpha_t - \alpha_{\min}}{\alpha_{\max} - \alpha_{\min}}.
\end{equation}

\noindent\emph{Criticality scoring and frame weighting.}
After jointly normalizing the two variables, we formally define the criticality score as:
\begin{equation}
  s_t = \max(\hat{v}_t,\, \hat{\alpha}_t),
  \label{eq:criticality}
\end{equation}
and the per-frame optimization weight is derived via a temperature-scaled softmax to prioritize critical timesteps during optimization:
\begin{equation}
  w_t = \exp(s_t/\tau) \Big/ \sum\nolimits_{t'=1}^{T} \exp(s_{t'}/\tau),
  \label{eq:weight}
\end{equation}
where $\tau > 0$ controls the concentration.
High $\hat{v}_t$ indicates rapid perceptual change; high $\hat{\alpha}_t$ signals an abrupt shift in the rate of change—hallmarks of behaviorally critical moments.
This instantiates $\mathbb{E}_{t \sim \mathcal{T}}$ in Eq.~\eqref{eq:objective} as a dynamics-weighted sum over the episode.

\noindent\textbf{Vertex-based texture parameterization.}
Direct optimization in the high-dimensional pixel space of $\mathbf{T}_{adv} \in \mathbb{R}^{H_t \times W_t \times 3}$ often tends to produce highly irregular perturbations that overfit to the specific model being attacked, resulting in poor cross-model transferability~\cite{huang2024towards}.
To mitigate this, we effectively leverage Nvdiffrast to reparameterize $\mathbf{T}_{adv}$ as $N_v$ per-vertex color attributes $\mathbf{c} \in \mathbb{R}^{N_v \times 3}$, with the texture map recovered through barycentric interpolation over the fixed mesh geometry $(\mathbf{V}, \mathbf{F})$ defined in Sec.~\ref{sec:formula}:
\begin{equation}
  \mathbf{T}_{adv} = \phi(\mathbf{c}).
  \label{eq:param}
\end{equation}
Since $N_v \ll H_t \times W_t$, this parameterization implicitly restricts the perturbation to a smooth, low-rank manifold defined by the mesh geometry, thereby effectively reducing the search space and consistently yielding more transferable adversarial textures.

\noindent\textbf{Untargeted attack.}
The untargeted objective degrades overall task success of the VLA model, without prescribing a specific failure mode.
Given the reference action $\mathbf{a}^*_{t,m} = \pi(\mathbf{O}_{t,m}(\mathcal{M}), l)$ produced by the model on the clean object under the $m$-th sampled view, we maximize the weighted action deviation averaged over views:
\begin{equation}
  \mathbf{c}^* = \underset{\mathbf{c}}{\arg\max}
  \sum_{t=1}^{T} w_t \cdot \frac{1}{M}\sum_{m=1}^{M}
  \bigl\|\pi\!\left(\mathbf{O}_{t,m}(\mathcal{M}_{adv}),\, l\right) - \mathbf{a}^*_{t,m}\bigr\|_2.
  \label{eq:untargeted}
\end{equation}

\noindent\textbf{Targeted attack.}
Targeted manipulation attacks go further by steering the model toward a prescribed erroneous trajectory, enabling deliberate behavioral hijacking.
Let $\mathbf{a}^{\mathrm{tgt}}_t$ denote the target action at timestep $t$.
The objective minimizes the deviation of the model's output from this trajectory across all sampled views:
\begin{equation}
  \mathbf{c}^* = \underset{\mathbf{c}}{\arg\min}
  \sum_{t=1}^{T} w_t \cdot \frac{1}{M}\sum_{m=1}^{M}
  \bigl\|\pi\!\left(\mathbf{O}_{t,m}(\mathcal{M}_{adv}),\, l\right) - \mathbf{a}^{\mathrm{tgt}}_t\bigr\|_2.
  \label{eq:targeted}
\end{equation}
In our implementation, $\mathbf{a}^{\mathrm{tgt}}_t$ is defined as semantically plausible, time-varying actions that redirect the gripper toward an alternative target at each timestep, inducing controlled task deviation.

\subsection{Physical Attack}
\label{method:physical_attack}

Adversarial textures optimized purely in the digital domain may degrade in effectiveness when applied to physical objects, due to real-world variations such as viewpoint shifts, lighting changes, and imaging artifacts.
To bridge this gap, we incorporate the Expectation over Transformations (EoT)~\cite{athalye2018synthesizing} strategy during optimization.
Rather than optimizing for a single rendered observation, EoT samples a stochastic transformation $g \sim \mathcal{T}$ at each step, covering both 3D variations (object pose perturbations, viewpoint shifts, distance changes) and 2D image-level augmentations (brightness, contrast, and blur).
The rendered observation under EoT is thus:
\begin{equation}
  \hat{\mathbf{O}}_t(\mathcal{M}_{adv}) = g\!\left(\mathbf{O}_t(\mathcal{M}_{adv})\right), \quad g \sim \mathcal{T},
  \label{eq:eot}
\end{equation}
and $\hat{\mathbf{O}}_t$ replaces $\mathbf{O}_t$ in Eqs.~\eqref{eq:untargeted} and~\eqref{eq:targeted} during training.
This encourages the adversarial texture to remain effective across diverse and unseen physical rendering conditions, improving its sim-to-real transferability. EoT complements the multi-view optimization in Sec.~\ref{sec:formula}: multi-view optimization improves robustness to viewpoint changes in simulation, whereas EoT further bridges the gap to physical deployment by modeling broader real-world transformations. Detailed analyses and pseudocode are in Appendices \textcolor[HTML]{974685}{A} and \textcolor[HTML]{974685}{B}.

\begin{table}[t]
  \centering
  \caption{Task failure rates (\%) comparison across color variants and Gaussian perturbation on four LIBERO task suites.}
  \label{tab:color_task_comparison}
  \footnotesize
  \setlength{\tabcolsep}{5pt}
  \resizebox{0.95\linewidth}{!}{
  \begin{tabular}{llcccc}
    \toprule
    Model & Variant & Spatial & Object & Goal & Long \\
    \midrule
    \multirow{5}{*}{OpenVLA}
      & Green    & \scoregain{20.2}{4.6} & \scoregain{17.8}{6.0} & \scoregain{27.4}{3.8} & \scoregain{48.8}{3.4} \\
      & Red      & \scoregain{23.2}{7.6} & \scoregain{18.6}{6.8} & \scoregain{26.6}{3.0} & \scoregain{50.6}{5.2} \\
      & Yellow   & \scoregain{18.8}{3.2} & \scoregain{15.4}{3.6} & \scoregain{25.2}{1.6} & \scoregain{48.4}{3.0} \\
      & Blue     & \scoregain{24.4}{8.8} & \scoregain{19.2}{7.4} & \scoregain{27.6}{4.0} & \scoregain{49.2}{3.8} \\
      & Gaussian & \scoregain{24.6}{9.0} & \scoregain{18.8}{7.0} & \scoregain{28.8}{5.2} & \scoregain{52.2}{6.8} \\
    \midrule
    \multirow{5}{*}{OpenVLA-OFT}
      & Green    & \scoregain{ 5.8}{2.0} & \scoregain{ 5.4}{3.7} & \scoregain{ 7.2}{3.4} & \scoregain{11.4}{2.1} \\
      & Red      & \scoregain{ 6.8}{3.0} & \scoregain{ 6.8}{5.1} & \scoregain{ 8.6}{4.8} & \scoregain{12.8}{3.5} \\
      & Yellow   & \scoregain{ 4.6}{0.8} & \scoregain{ 4.8}{3.1} & \scoregain{ 6.8}{3.0} & \scoregain{10.8}{1.5} \\
      & Blue     & \scoregain{ 6.2}{2.4} & \scoregain{ 5.9}{4.2} & \scoregain{ 8.8}{5.0} & \scoregain{13.9}{4.6} \\
      & Gaussian & \scoregain{ 8.8}{5.0} & \scoregain{ 3.6}{1.9} & \scoregain{ 5.4}{1.6} &\scoregain{ 10.9}{1.6}\\
    \bottomrule
  \end{tabular}}
\end{table}
\section{Experiments}
\subsection{Experimental Setup}

\noindent\textbf{Datasets \& Threat Models.} We conduct our experiments on the LIBERO~\cite{liu2023libero} benchmark in simulation environments, which provides a standardized testbed for evaluating VLA models across four task categories: Spatial, Object, Goal, and Long-horizon tasks. These tasks cover progressively more challenging  manipulation scenarios,ranging from simple spatial reasoning to multi-step planning. We evaluate four representative VLA models, OpenVLA, OpenVLA-OFT, $\pi_0$, and $\pi_{0.5}$, and report results on all LIBERO task variants.

\noindent\textbf{Baselines.} We use three variants with \methodname{}: (1) \textit{Single-frame}: perturbs the adversarial texture using only a single observation frame~\cite{xiao2019meshadv}; (2) \textit{Vertex Param.}: removes temporal consistency and performs only vertex-based adversarial texture optimization; and (3) \textit{Tex+Temp.}: introduces temporal consistency to the texture-only variant. Additionally, following~\cite{dong2022viewfool,ruan2023towards}, we establish two potential baselines settings : the clean setting without perturbation (\textit{No Attack}) and random Gaussian perturbations applied to the object texture (\textit{Gaussian}). In Sec.~\ref{sec:additional}, we further compare \methodname{} against 2D patch-based adversarial attacks~\cite{wang2025exploring} in terms of robustness under digital geometric variations and physical-world deployment.


\begin{figure}[!t]
  \centering
  \includegraphics[width=0.9\linewidth]{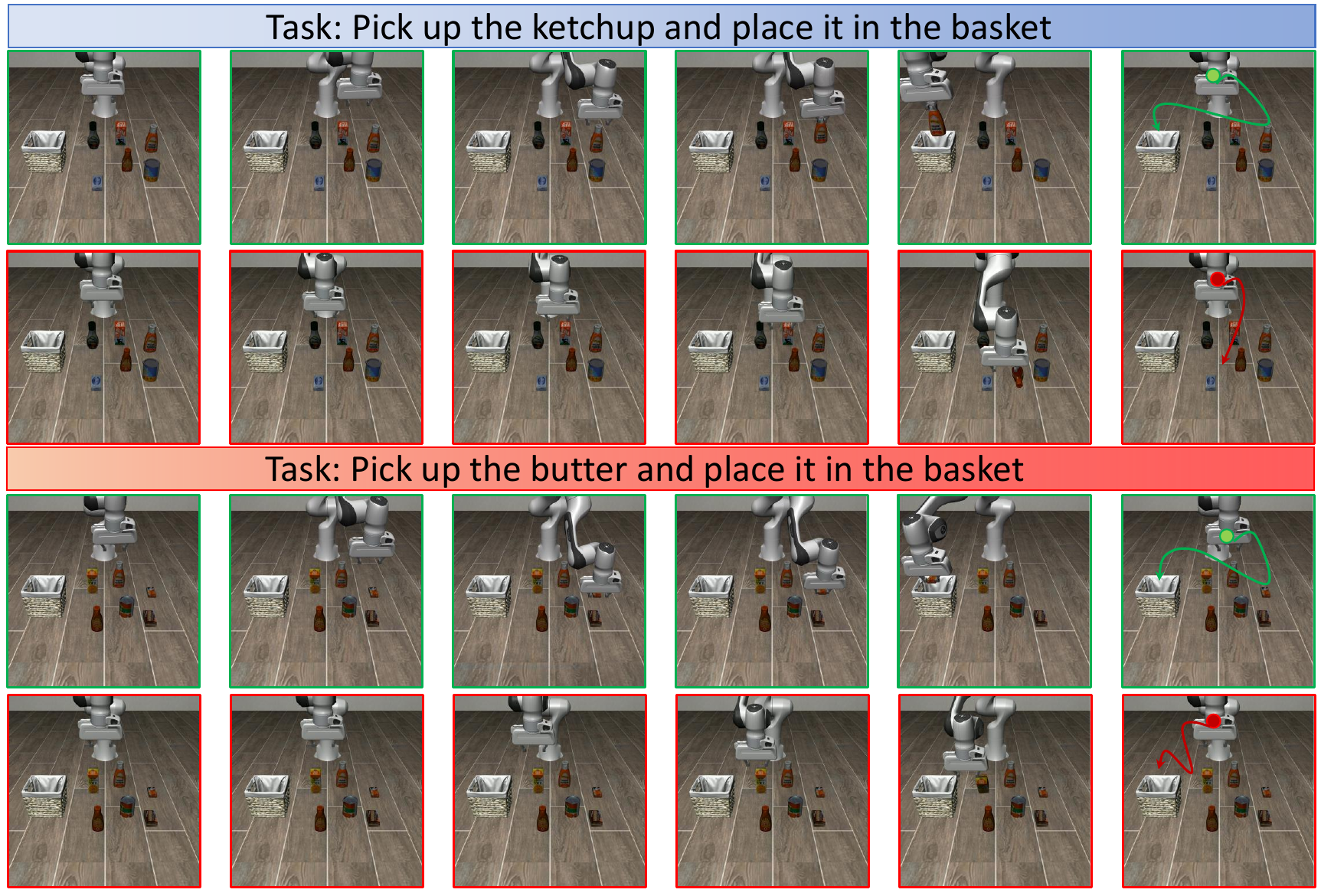}
  \caption{Qualitative results of \methodname{} on manipulation tasks. For each task, the green row shows the clean rollout,whereas the red row shows the adversarial rollout under \methodname{}.}
  \label{fig:attack_visual}
\end{figure}

\begin{figure}[t]
  \centering
  \includegraphics[width=\linewidth]{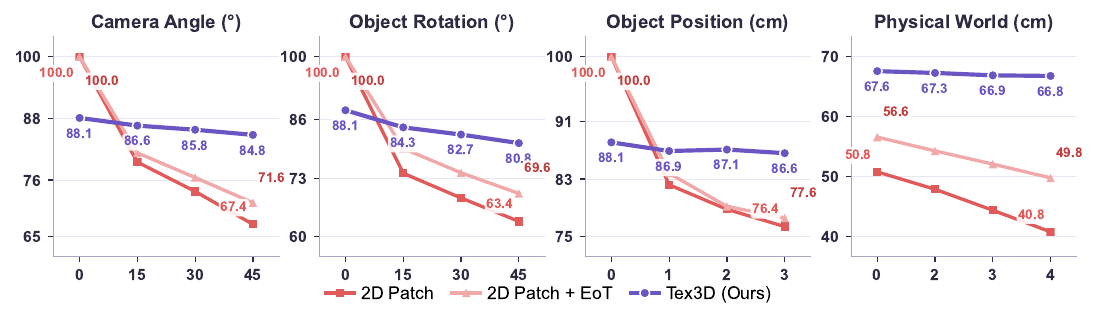}
  \caption{Robustness comparison of \methodname{} and 2D patch-based baselines under varying camera angles, object rotations, object positions (digital simulation), and position offsets in the physical world. Task failure rate (\%, $\uparrow$) is reported.}
  \label{fig:robustness}
\end{figure}
\noindent\textbf{Evaluation Metric \& Details.}
We evaluate \methodname{} following the standard protocol adopted in the LIBERO benchmark. Each task is executed for 50 independent trials, and performance is measured by \textit{Task Failure Rate} (FR), defined as the proportion of failed task completions over all trials. For each benchmark suite, we report the average FR across tasks to reflect overall attack effectiveness and generalization. Unless otherwise specified, all experiments are conducted on a single NVIDIA A100 GPU with 80GB memory.

\noindent\textbf{Physical Experiment Setting.} For real-world experiments, we employ a robotic platform built around a Franka Emika Panda manipulator, controlled via 7-DoF end-effector delta commands. Observations are obtained from a single monocular RGB camera, and our method operates solely on raw RGB inputs without requiring additional sensory modalities. To facilitate systematic quantitative evaluation, we utilize 3D printing to fabricate both the objects used in pick-and-place tasks and the corresponding adversarial variants. All tasks are repeated for 100 independent trials each. Further implementation details and setups are provided in Appendix \textcolor[HTML]{974685}{G}.
\subsection{Main Results}

\begin{figure}[t]
  \centering
  \includegraphics[width=0.95\linewidth]{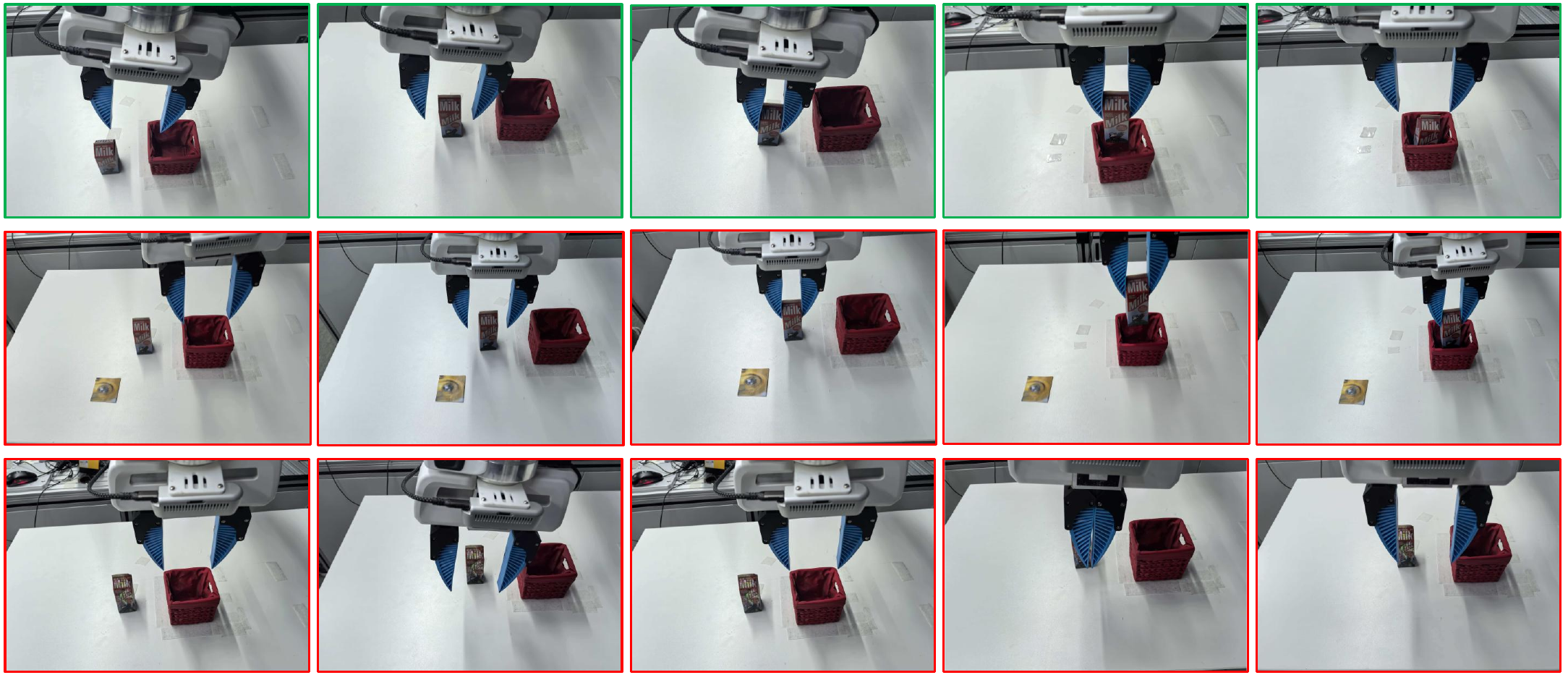}
  \caption{Physical-world qualitative comparison. The first row: clean samples, the second row :results under 2D patch-based attacks, and the third row: results under \methodname{}.}
  \label{fig:physical}
\end{figure}
\noindent\textbf{Overall Attack Effectiveness.} Tab.~\ref{tab:main} quantitatively summarizes the overall effectiveness of \methodname{} on LIBERO tasks. \methodname{} consistently yields the highest task failure rates across all four victim models, all four task variants, and both untargeted and targeted settings. On OpenVLA, the average failure rate rises sharply from 24.1\% to 88.1\% and 90.5\% under untargeted and targeted attacks, respectively, while OpenVLA-OFT increases from 4.7\% to 76.0\% and 79.3\%. Similar trends are observed on $\pi_0$, whose average failure rate rises from 4.6\% to 71.8\% and 73.3\%, and on $\pi_{0.5}$, where it increases from 2.8\% to 69.3\% and 71.2\%. For targeted attacks, following the evaluation setting in~\cite{wang2025exploring}, Tab.~\ref{tab:zero_action_metrics} further shows that the attacked actions remain close to the prescribed target actions, with average L1 distances of 0.0176, 0.0237, 0.0305, and 0.0372 on OpenVLA, OpenVLA-OFT, $\pi_0$, and $\pi_{0.5}$, respectively. The most pronounced increase appears on the Spatial task, where \methodname{} raises the failure rate from 15.6\% to 95.8\%/96.7\% on OpenVLA, and from 3.5\% to 74.9\%/75.9\% on $\pi_0$. These results indicate that the strong attack performance does not come from simple noise injection or single-frame perturbation, but from jointly optimizing object-level adversarial textures with temporal consistency. This principled design enables \methodname{} to produce stronger and more stable attacks than all baselines and ablations across diverse manipulation tasks.


\noindent\textbf{Cross-Model Transferability.} Tab.~\ref{tab:analysis}~(bottom) reports cross-model transfer results to evaluate whether adversarial textures optimized on one VLA generalize to unseen victim models. Strong transfer appears in both same- and cross-family settings in practice. Within the same family, attacks transfer strongly between OpenVLA and OpenVLA-OFT, yielding 61.5\%--70.6\% failure rates in the OpenVLA $\to$ OpenVLA-OFT direction and 70.2\%--75.7\% in the reverse direction; transfer between $\pi_0$ and $\pi_{0.5}$ is likewise consistently high, at 49.2\%--58.4\% and 55.8\%--63.7\%, respectively. Across families, attacks crafted on OpenVLA transfer effectively to $\pi_0$ and $\pi_{0.5}$, reaching 34.5\%--44.1\% and 27.8\%--36.5\%, while textures optimized on $\pi_0$ remain effective on OpenVLA and OpenVLA-OFT, achieving 44.8\%--61.5\% and 37.2\%--48.4\%, respectively. Taken together, these results show that \methodname{} learns robust object-level adversarial patterns rather than overfitting to a specific source model, making the attack also effective under diverse source-target model pairings.

\noindent\textbf{Ruling Out Non-adversarial Factors.} To verify the elevated failure rates in Tab.~\ref{tab:main} arise from adversarial optimization rather than color variation or rendering artifacts, two control studies are performed. First, Fig.~\ref{fig:first} and Tab.~\ref{tab:color_task_comparison} show that neither color changes nor random Gaussian perturbations cause comparably severe failures. For instance, on the OpenVLA Spatial task, these controls remain within 18.8\%--24.6\%, far below the 95.8\%/96.7\% achieved by \methodname{} under untargeted/targeted attacks. Second, Tab.~\ref{tab:analysis}~(top) shows that the MuJoCo+nvdiffrast cross-rendering pipeline introduces negligible discrepancies, with at most 1.2 points difference across all four task suites. Together, these results confirm that the high failure rates are primarily induced by the adversarial textures optimized by \methodname{}, rather than ordinary color shifts or rendering distortion. 

\subsection{Additional Analysis}
\label{sec:additional}

\begin{table}[t]
  \centering
  \caption{L1 distance ($\downarrow$) between the attacked action and target action across task suites on four victim VLA models.}
  \label{tab:zero_action_metrics}
  \footnotesize
  \setlength{\tabcolsep}{7pt}
  \begin{tabular}{lccccc}
    \toprule
    Model & Spatial & Object & Goal & Long & Avg. \\
    \midrule
    OpenVLA & 0.0137 & 0.0205 & 0.0185 & 0.0176 & 0.0176 \\
    OpenVLA-OFT & 0.0214 & 0.0281 & 0.0241 & 0.0212 & 0.0237 \\
    \midrule
    $\pi_0$ & 0.0284 & 0.0357 & 0.0312 & 0.0268 & 0.0305 \\
    $\pi_{0.5}$ & 0.0349 & 0.0423 & 0.0386 & 0.0331 & 0.0372 \\
    \bottomrule
  \end{tabular}
\end{table}

\begin{table}[!t]
  \centering
  \caption{Representative trajectory-aware dynamics-guided frame weights along a sample manipulation trajectory.}
  \label{tab:keyframe}
  \footnotesize
  \setlength{\tabcolsep}{4pt}
  \begin{tabular}{lccccc}
    \toprule
    Stage & T1 Initial & T2 Reach & T3 Grasp & T4 Transfer & T5 Put down \\
    \midrule
    Step & 10 & 50 & 100 & 150 & 195 \\
    Weight & 0.20 & 0.37 & 0.90 & 0.32 & 0.20 \\
    \bottomrule
  \end{tabular}
\end{table}
\begin{figure}[t]
  \centering
  \includegraphics[width=0.95\linewidth]{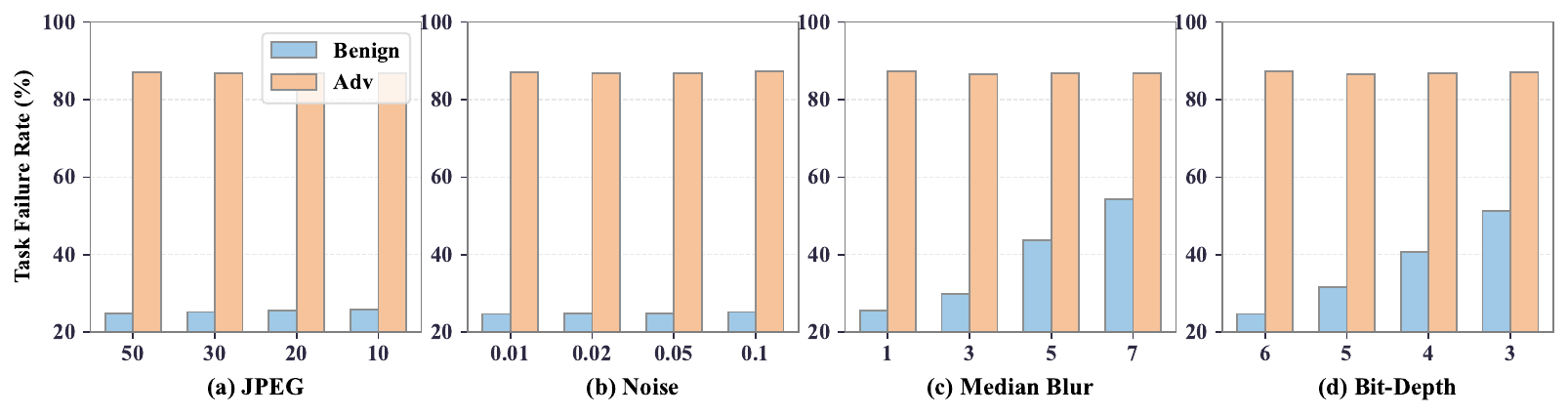}
  \caption{Impact of input-space defenses on \methodname{}.}
  \label{fig:defense}
\end{figure}

\begin{figure}[t]
  \centering
  \includegraphics[width=0.99\linewidth]{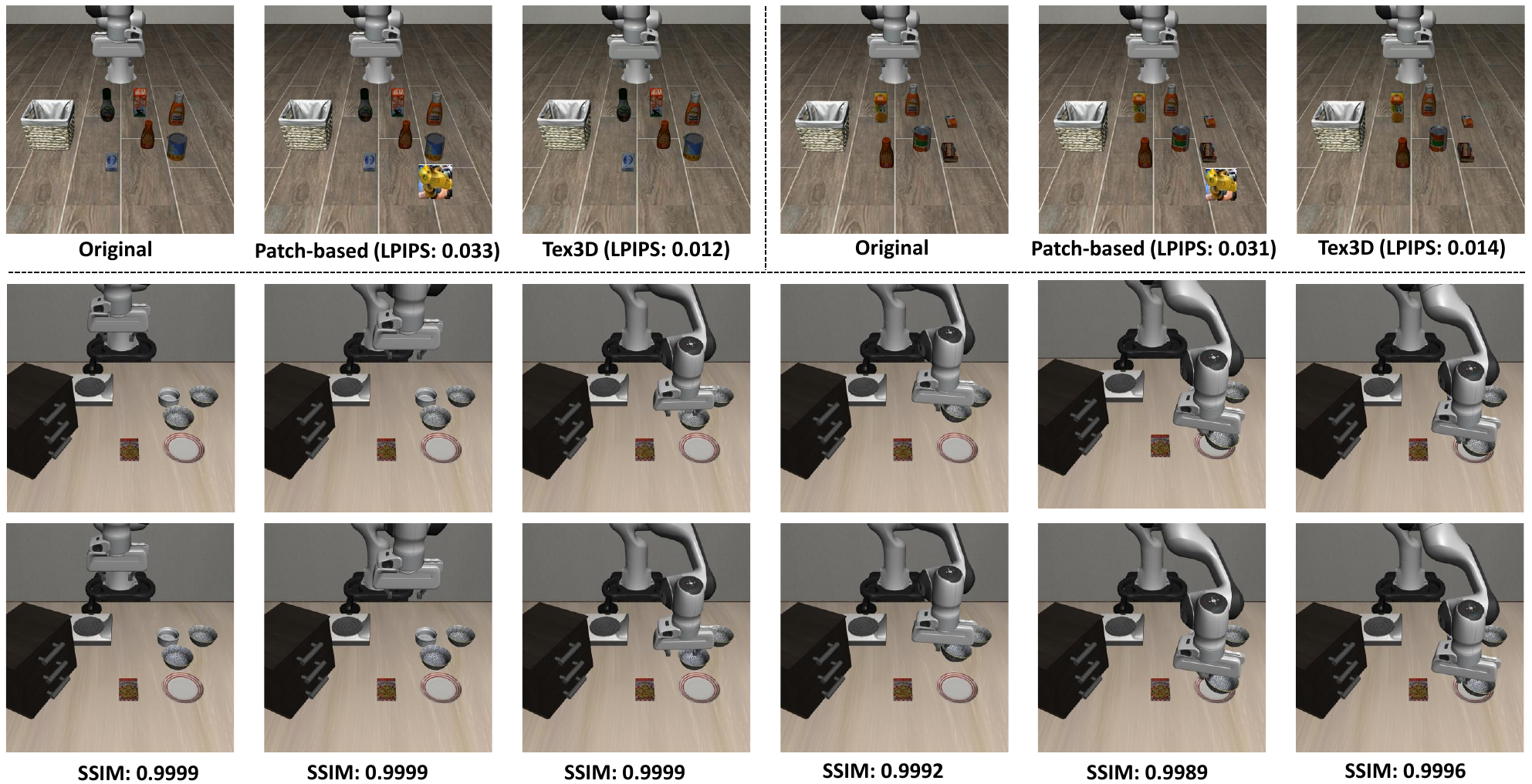}
  \caption{First column: qualitative perceptual-similarity direct comparison between \methodname{} and 2D patch-based attacks. Second and third columns: rendering-consistency comparison at the same optimization step between the original renderer (MuJoCo) and our dual-renderer pipeline (Ours).}
  \label{fig:vspatch}
\end{figure}

\noindent\textbf{Robustness against 2D Patch Attacks.} To further systematically examine whether object-level adversarial textures provide better robustness than conventional 2D patch-based attacks~\cite{wang2025exploring}, additional comparisons are conducted in both digital and physical settings. Fig.~\ref{fig:robustness} shows that 2D patch-based attacks are highly brittle under \textbf{digital} geometric variations: their failure rate drops sharply from 100\% to 67.4\% under camera-angle perturbations and to 63.4\% under object rotations. In contrast, \methodname{} consistently remains much more stable, staying around 80.8\%--88.1\% across all three digital variations. A similar advantage is preserved in the \textbf{physical} world, where \methodname{} maintains substantially higher failure rates (66.8\%--67.6\%) than both 2D patch (40.8\%--50.8\%) and 2D patch+EoT (49.8\%--56.6\%). Fig.~\ref{fig:physical} further provides qualitative evidence that 2D patches are more sensitive to viewpoint changes and placement mismatch, whereas the object-level adversarial textures generated by \methodname{} preserve their attack effect after physical deployment. In addition, the first row of Fig.~\ref{fig:vspatch} shows that \methodname{} also achieves better visual quality, reflected by lower LPIPS than the 2D patch-based baseline. Overall, these results indicate that optimizing textures directly on the target object yields attacks that are not only more transferable across geometric variations, but also more robust and consistently visually natural in real-world deployment scenarios, even under diverse and challenging real-world environments.

\noindent\textbf{Robustness Against Common Defenses.} Beyond the above geometric variations such as rotations and viewpoint changes, we further study the robustness of \methodname{} against common defenses~\cite{dziugaite2016study,zhang2019defending,xu2017feature}. Fig.~\ref{fig:defense} shows that standard input-space defenses, including JPEG compression, additive noise, median blur, and bit-depth reduction, have little effect on mitigating our attack. Across all defense settings, the adversarial task failure rate under \methodname{} remains nearly unchanged, staying around 86.6\%--87.3\%. These results suggest that \methodname{} preserves strong attack effectiveness even under common defenses, further demonstrating its robustness.

\noindent\textbf{Rendering Consistency and Attack Trajectory.} To further empirically verify that the proposed attack does not rely on renderer mismatch, the second and third columns of Fig.~\ref{fig:vspatch} compare same-step renderings from the original MuJoCo pipeline and our dual-renderer pipeline. The resulting SSIM scores are consistently close to 1, indicating that the dual-renderer composition remains highly consistent with the original simulator rendering while still enabling gradient-based optimization. At the same time, Fig.~\ref{fig:attack_visual} shows that these visually consistent renderings can nevertheless induce substantial behavioral deviations over time: compared with the clean rollout, the adversarial trajectory gradually drifts away from the intended manipulation path and eventually leads to clear task failure. Together, these results show that the effectiveness of \methodname{} comes from adversarial texture optimization rather than rendering artifacts, while our dual-renderer design preserves simulator fidelity and still produces consistently strong trajectory-level attack effects.

\noindent\textbf{Representative Keyframe Weights.} To further illustrate how TAAO allocates optimization effort over time, Tab.~\ref{tab:keyframe} reports representative frame weights along a manipulation trajectory. The weights remain relatively low at stable stages such as initialization and put-down, become moderately larger during reaching and transfer, and peak sharply at the particularly critical grasping stage. In this example, grasping receives the highest weight, indicating that TAAO concentrates adversarial optimization on the most decision-sensitive moment of the task. This behavior is consistent with the intuition that errors near grasp formation are more likely to propagate to subsequent stages and cause final task failure. An ablation study on keyframe inversion is provided in Appendix \textcolor[HTML]{974685}{C}.

\begin{table}[t]
  \centering
  \caption{Ablation study on FBD and TAAO components of \methodname{}. Fail. Rate: Task Failure Rate on OpenVLA (\%, $\uparrow$); Time: wall-clock time per gradient optimization step (s, $\downarrow$).}
  \label{tab:ablation}
  \footnotesize
  \setlength{\tabcolsep}{4pt}
  \begin{tabular}{ccc|ccc|cc}
    \toprule
    \multicolumn{3}{c|}{FBD} & \multicolumn{3}{c|}{TAAO} & Fail. Rate & Time \\
    \cmidrule(lr){1-3}\cmidrule(lr){4-6}
    MVP & Light. & Decouple & Rand. & Unif. & Dyn. & (\%) & (s) \\
    \midrule
    \multicolumn{8}{l}{\textit{FBD Component Ablation}} \\
    \midrule
    --        & \ding{51} & \ding{51} & -- & -- & \ding{51} & 65.8 & {\raise.17ex\hbox{$\scriptstyle\sim$}}7.2 \\
    \ding{51} & --        & \ding{51} & -- & -- & \ding{51} & 76.8 & {\raise.17ex\hbox{$\scriptstyle\sim$}}7.2 \\
    \ding{51} & \ding{51} & --        & -- & -- & \ding{51} & 84.6 & {\raise.17ex\hbox{$\scriptstyle\sim$}}24.8 \\
    \midrule
    \multicolumn{8}{l}{\textit{TAAO Weighting Strategy Ablation}} \\
    \midrule
    \ding{51} & \ding{51} & \ding{51} & \ding{51} & --        & --        & 73.7 & {\raise.17ex\hbox{$\scriptstyle\sim$}}7.2 \\
    \ding{51} & \ding{51} & \ding{51} & --        & \ding{51} & --        & 82.1 & {\raise.17ex\hbox{$\scriptstyle\sim$}}7.2 \\
    \rowcolor{MethodTint}
    \ding{51} & \ding{51} & \ding{51} & --        & --        & \ding{51} & 88.1 & {\raise.17ex\hbox{$\scriptstyle\sim$}}7.2 \\
    \bottomrule
  \end{tabular}
\end{table}

\begin{figure}[!t]
  \centering
  \includegraphics[width=\linewidth]{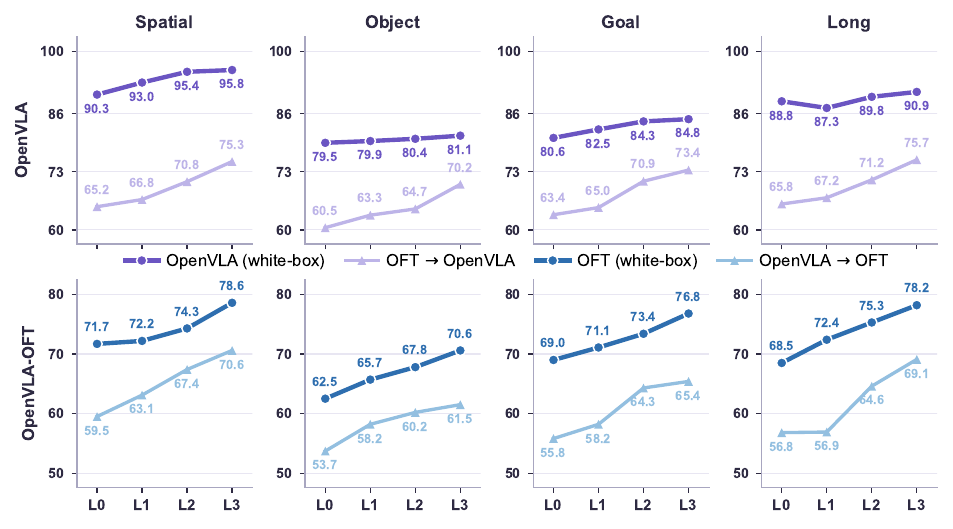}
  \caption{\methodname{} performance across four perturbation levels. Each subplot shows task failure rates (\%) from L0 to L3 settings, where L0 denotes the naturalness-constrained setting and L1--L3 progressively increase the perturbation budget.}
  \label{fig:libero_adv_curves}
\end{figure}
\subsection{ Ablation Studies}

\noindent\textbf{Effect of FBD and TAAO Design.} Tab.~\ref{tab:ablation} studies which components drive the effectiveness and efficiency of \methodname{}. Removing \textit{Decouple}, i.e., rendering the whole scene in nvdiffrast instead of separating foreground and background, reduces task failure rate from 88.1\% to 84.6\% and increases per-step optimization time from \raise.17ex\hbox{$\scriptstyle\sim$}7.2s to \raise.17ex\hbox{$\scriptstyle\sim$}24.8s. Removing MVP or Lighting further lowers the task failure rate to 65.8\% and 76.8\%, respectively, confirming the importance of cross-renderer alignment and lighting modeling. On the TAAO side, dynamics-guided weighting achieves the highest task failure rate, outperforming random weighting (73.7\%) and uniform weighting (82.1\%). Overall, dual-renderer decoupling provides a favorable effectiveness-efficiency trade-off, while dynamics-guided weighting is the most effective choice for trajectory-aware optimization,consistently yielding better performance across tasks.

\begin{figure}[t]
  \centering
  \includegraphics[width=0.95\linewidth]{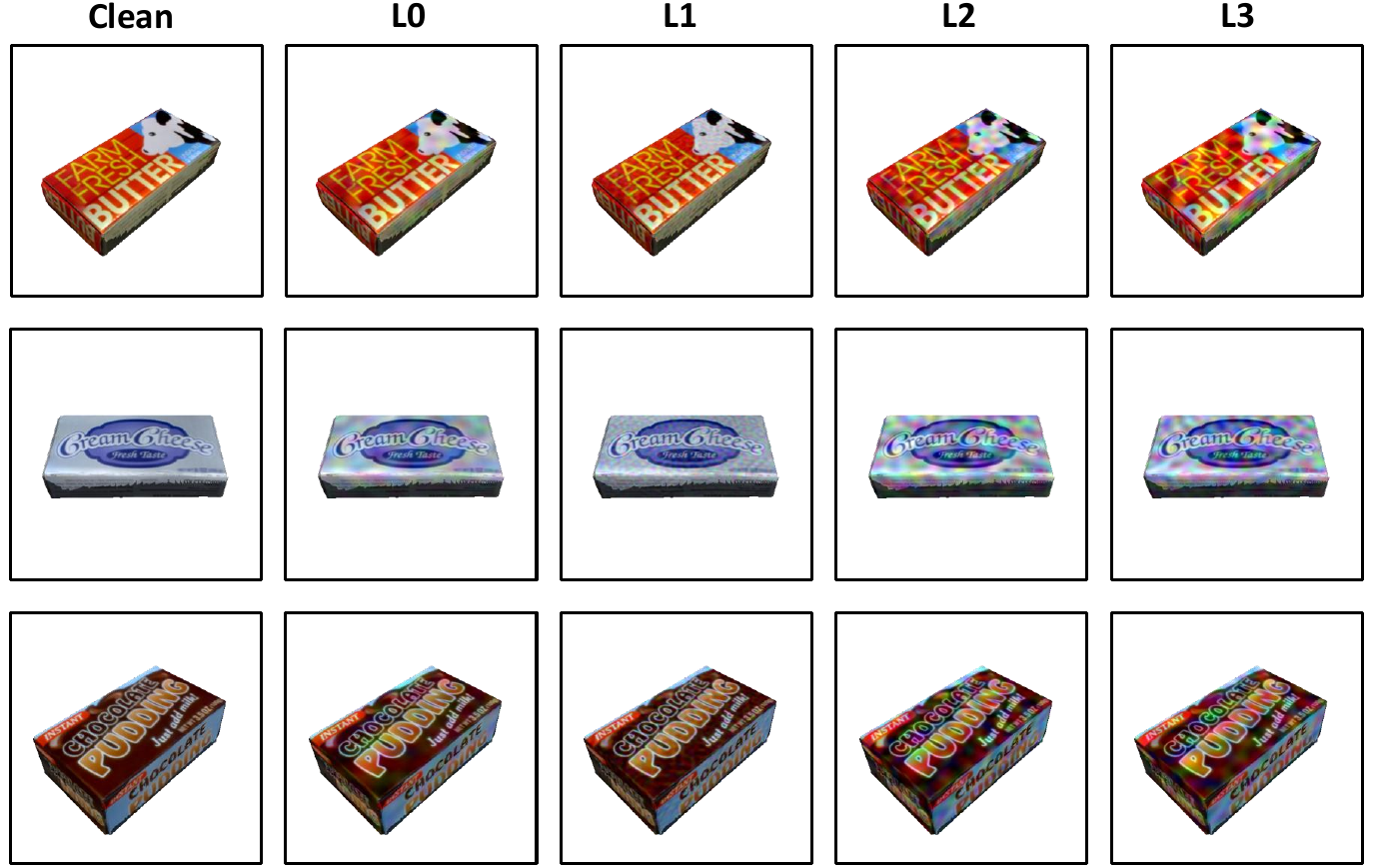}
  \caption{The qualitative visual comparison under different perturbation budgets and the clean reference sample image.}
  \label{fig:level}
\end{figure}

\noindent\textbf{Effect of Perturbation Level.} To study how perturbation strength affects attack performance, we vary the perturbation level from L0 to L3 and report both white-box and cross-model transfer results in Fig.~\ref{fig:libero_adv_curves}. Fig.~\ref{fig:level} further provides a qualitative comparison of adversarial textures under different budgets. Here, L0 denotes the naturalness-constrained setting with $\varepsilon=64/255$ and an additional MSE loss between adversarial and clean samples, while L1, L2, and L3 correspond to $\varepsilon=16/255$, $32/255$, and $64/255$, respectively. Overall, all perturbation levels substantially increase task failure rates across all four LIBERO task suites for both victim models. A clear trend shows that stronger perturbations result in higher failure rates, especially from L1 to L3, in both white-box and transfer settings. Meanwhile, L0 yields slightly lower failure rates than unconstrained levels, indicating a trade-off between stealthiness and attack strength. Despite this, L0 remains highly effective, suggesting that visually natural perturbations can still induce severe failures. These results demonstrate that the proposed attack is robust across a wide range of perturbation budgets, with stronger perturbations further increasing failure rates and improving transferability.

\section{Discussion and Conclusion}
Figs.~\ref{fig:libero_adv_curves} and~\ref{fig:level} reveal a noteworthy phenomenon: even nearly imperceptible perturbations under the L0 and L1 settings can already cause sharp increases in VLA failure rates, with OpenVLA still exceeding 90\% on the Spatial suite under strong attack. This fragile robustness may stem from the current data regime of VLA systems, as existing robot training corpora and evaluation benchmarks are still largely built in relatively clean and idealized environments~\cite{fei2025libero} with limited coverage of subtle appearance shifts and adversarial visual variations. As a result, even small perturbations can induce substantial distribution shifts in the perception module. Our findings therefore highlight the urgent need to rethink VLA training for meaningful robustness against adversarial attacks. Such robustness may potentially be improved in two directions: \textit{(1)} incorporating more diverse and realistic robotic scenes at the visual level; and \textit{(2)} enforcing stronger structural constraints over joint dependencies at the action level to filter out implausible or harmful actions.

In summary, this paper presents \methodname{}, the first framework for end-to-end optimization of adversarial 3D textures directly in the VLA simulation environment. By introducing FBD, we establish a differentiable optimization path from VLA objectives back to object appearance while preserving rendering consistency with the original simulator. Built on this, TAAO emphasizes behaviorally critical frames to sustain attack effectiveness over long-horizon manipulation trajectories. Extensive experiments in both simulation and physical settings across multiple VLA models and LIBERO task suites show that \methodname{} consistently achieves high task failure rates under both untargeted and targeted settings, transfers effectively across model architectures, and remains robust under geometric variations and common defenses. Taken together, these results show that object-level adversarial textures constitute a practical and highly effective attack surface for embodied agents,and underscore the need for more systematic robustness evaluation and more comprehensive training under persistent, physically grounded adversarial interactions in future real-world deployment settings and safety-critical scenarios, particularly in complex environments.

\clearpage
\bibliographystyle{ACM-Reference-Format}
\bibliography{sample-base}

@article{brohan2022rt,
  title={Rt-1: Robotics transformer for real-world control at scale},
  author={Brohan, Anthony and Brown, Noah and Carbajal, Justice and Chebotar, Yevgen and Dabis, Joseph and Finn, Chelsea and Gopalakrishnan, Keerthana and Hausman, Karol and Herzog, Alex and Hsu, Jasmine and others},
  journal={arXiv preprint arXiv:2212.06817},
  year={2022}
}

@inproceedings{zitkovich2023rt,
  title={Rt-2: Vision-language-action models transfer web knowledge to robotic control},
  author={Zitkovich, Brianna and Yu, Tianhe and Xu, Sichun and Xu, Peng and Xiao, Ted and Xia, Fei and Wu, Jialin and Wohlhart, Paul and Welker, Stefan and Wahid, Ayzaan and others},
  booktitle={Conference on Robot Learning},
  pages={2165--2183},
  year={2023},
  organization={PMLR}
}

@article{team2024octo,
  title={Octo: An open-source generalist robot policy},
  author={Team, Octo Model and Ghosh, Dibya and Walke, Homer and Pertsch, Karl and Black, Kevin and Mees, Oier and Dasari, Sudeep and Hejna, Joey and Kreiman, Tobias and Xu, Charles and others},
  journal={arXiv preprint arXiv:2405.12213},
  year={2024}
}

@Article{Kim2024OpenVLAAO,
  title={Openvla: An open-source vision-language-action model},
  author={Kim, Moo Jin and Pertsch, Karl and Karamcheti, Siddharth and Xiao, Ted and Balakrishna, Ashwin and Nair, Suraj and Rafailov, Rafael and Foster, Ethan and Lam, Grace and Sanketi, Pannag and others},
  journal={arXiv preprint arXiv:2406.09246},
  year={2024}
}

@article{liu2024rdt,
  title={Rdt-1b: a diffusion foundation model for bimanual manipulation},
  author={Liu, Songming and Wu, Lingxuan and Li, Bangguo and Tan, Hengkai and Chen, Huayu and Wang, Zhengyi and Xu, Ke and Su, Hang and Zhu, Jun},
  journal={arXiv preprint arXiv:2410.07864},
  year={2024}
}

@article{black2024pi_0,
  title={$pi\_0 $: A Vision-Language-Action Flow Model for General Robot Control},
  author={Black, Kevin and Brown, Noah and Driess, Danny and Esmail, Adnan and Equi, Michael and Finn, Chelsea and Fusai, Niccolo and Groom, Lachy and Hausman, Karol and Ichter, Brian and others},
  journal={arXiv preprint arXiv:2410.24164},
  year={2024}
}

@article{wen2025tinyvla,
  title={Tinyvla: Towards fast, data-efficient vision-language-action models for robotic manipulation},
  author={Wen, Junjie and Zhu, Yichen and Li, Jinming and Zhu, Minjie and Tang, Zhibin and Wu, Kun and Xu, Zhiyuan and Liu, Ning and Cheng, Ran and Shen, Chaomin and others},
  journal={IEEE Robotics and Automation Letters},
  year={2025},
  publisher={IEEE}
}

@article{shukor2025smolvla,
  title={Smolvla: A vision-language-action model for affordable and efficient robotics},
  author={Shukor, Mustafa and Aubakirova, Dana and Capuano, Francesco and Kooijmans, Pepijn and Palma, Steven and Zouitine, Adil and Aractingi, Michel and Pascal, Caroline and Russi, Martino and Marafioti, Andres and others},
  journal={arXiv preprint arXiv:2506.01844},
  year={2025}
}

@article{liu2025hybridvla,
  title={Hybridvla: Collaborative diffusion and autoregression in a unified vision-language-action model},
  author={Liu, Jiaming and Chen, Hao and An, Pengju and Liu, Zhuoyang and Zhang, Renrui and Gu, Chenyang and Li, Xiaoqi and Guo, Ziyu and Chen, Sixiang and Liu, Mengzhen and others},
  journal={arXiv preprint arXiv:2503.10631},
  year={2025}
}

@inproceedings{karamcheti2024prismatic,
  title={Prismatic vlms: Investigating the design space of visually-conditioned language models},
  author={Karamcheti, Siddharth and Nair, Suraj and Balakrishna, Ashwin and Liang, Percy and Kollar, Thomas and Sadigh, Dorsa},
  booktitle={Forty-first International Conference on Machine Learning},
  year={2024}
}

@inproceedings{o2024open,
  title={Open x-embodiment: Robotic learning datasets and rt-x models: Open x-embodiment collaboration 0},
  author={O’Neill, Abby and Rehman, Abdul and Maddukuri, Abhiram and Gupta, Abhishek and Padalkar, Abhishek and Lee, Abraham and Pooley, Acorn and Gupta, Agrim and Mandlekar, Ajay and Jain, Ajinkya and others},
  booktitle={2024 IEEE International Conference on Robotics and Automation (ICRA)},
  pages={6892--6903},
  year={2024},
  organization={IEEE}
}

@article{zheng2024tracevla,
  title={Tracevla: Visual trace prompting enhances spatial-temporal awareness for generalist robotic policies},
  author={Zheng, Ruijie and Liang, Yongyuan and Huang, Shuaiyi and Gao, Jianfeng and Daum{\'e} III, Hal and Kolobov, Andrey and Huang, Furong and Yang, Jianwei},
  journal={arXiv preprint arXiv:2412.10345},
  year={2024}
}

@misc{qu2025spatialvlaexploringspatial,
      title={SpatialVLA: Exploring Spatial Representations for Visual-Language-Action Model}, 
      author={Qu, Delin and Song, Haoming and Chen, Qizhi and Yao, Yuanqi and Ye, Xinyi and Ding, Yan and Wang, Zhigang and Gu, JiaYuan and Zhao, Bin and Wang, Dong and Li, Xuelong},
      year={2025},
      eprint={2501.15830},
      archivePrefix={arXiv},
      primaryClass={cs.RO},
      url={https://arxiv.org/abs/2501.15830}, 
}

@article{li2024cogact,
  title={Cogact: A foundational vision-language-action model for synergizing cognition and action in robotic manipulation},
  author={Li, Qixiu and Liang, Yaobo and Wang, Zeyu and Luo, Lin and Chen, Xi and Liao, Mozheng and Wei, Fangyun and Deng, Yu and Xu, Sicheng and Zhang, Yizhong and others},
  journal={arXiv preprint arXiv:2411.19650},
  year={2024}
}

@misc{liang2025finetuningvisionlanguageactionmodels,
  title={Fine-tuning vision-language-action models: Optimizing speed and success},
  author={Kim, Moo Jin and Finn, Chelsea and Liang, Percy},
  journal={arXiv preprint arXiv:2502.19645},
  year={2025}
}

@inproceedings{liu2024exploring,
  title={Exploring the robustness of decision-level through adversarial attacks on llm-based embodied models},
  author={Liu, Shuyuan and Chen, Jiawei and Ruan, Shouwei and Su, Hang and Yin, Zhaoxia},
  booktitle={Proceedings of the 32nd ACM international conference on multimedia},
  pages={8120--8128},
  year={2024}
}

@misc{dupuy2025embodiedredteaming,
  title={Embodied red teaming for auditing robotic foundation models},
  author={Karnik, Sathwik and Hong, Zhang-Wei and Abhangi, Nishant and Lin, Yen-Chen and Wang, Tsun-Hsuan and Dupuy, Christophe and Gupta, Rahul and Agrawal, Pulkit},
  journal={arXiv preprint arXiv:2411.18676},
  year={2024}
}

@misc{yanSun,
      title={When Alignment Fails: Multimodal Adversarial Attacks on Vision-Language-Action Models},
      author={Yuping Yan and Yuhan Xie and Yixin Zhang and Lingjuan Lyu and Handing Wang and Yaochu Jin},
      year={Sun Nov 23 2025 07:21:40 GMT+0000 (Coordinated Universal Time)},
      eprint={2511.16203},
      archivePrefix={arXiv},
      primaryClass={cs.AI},
      url={https://arxiv.org/abs/2511.16203},
}

@article{jones2025adversarial,
  title={Adversarial attacks on robotic vision language action models},
  author={Jones, Eliot Krzysztof and Robey, Alexander and Zou, Andy and Ravichandran, Zachary and Pappas, George J and Hassani, Hamed and Fredrikson, Matt and Kolter, J Zico},
  journal={arXiv preprint arXiv:2506.03350},
  year={2025}
}

@inproceedings{wang2025exploring,
  title={Exploring the adversarial vulnerabilities of vision-language-action models in robotics},
  author={Wang, Taowen and Han, Cheng and Liang, James and Yang, Wenhao and Liu, Dongfang and Zhang, Luna Xinyu and Wang, Qifan and Luo, Jiebo and Tang, Ruixiang},
  booktitle={Proceedings of the IEEE/CVF International Conference on Computer Vision},
  pages={6948--6958},
  year={2025}
}

@article{liu2025eva,
  title={Eva-VLA: Evaluating Vision-Language-Action Models' Robustness Under Real-World Physical Variations},
  author={Liu, Hanqing and Long, Jiahuan and Wu, Junqi and Hou, Jiacheng and Tang, Huili and Jiang, Tingsong and Zhou, Weien and Yao, Wen},
  journal={arXiv preprint arXiv:2509.18953},
  year={2025}
}

@misc{luWed,
      title={When Robots Obey the Patch: Universal Transferable Patch Attacks on Vision-Language-Action Models},
      author={Hui Lu and Yi Yu and Yiming Yang and Chenyu Yi and Qixin Zhang and Bingquan Shen and Alex C. Kot and Xudong Jiang},
      year={Wed Nov 26 2025 09:16:32 GMT+0000 (Coordinated Universal Time)},
      eprint={2511.21192},
      archivePrefix={arXiv},
      primaryClass={cs.AI},
      url={https://arxiv.org/abs/2511.21192},
}

@article{jia2022physical,
  title={Physical adversarial attack on a robotic arm},
  author={Jia, Yifan and Poskitt, Christopher M and Sun, Jun and Chattopadhyay, Sudipta},
  journal={IEEE Robotics and Automation Letters},
  volume={7},
  number={4},
  pages={9334--9341},
  year={2022},
  publisher={IEEE}
}

@misc{zhangWed,
      title={Attention-Guided Patch-Wise Sparse Adversarial Attacks on Vision-Language-Action Models},
      author={Naifu Zhang and Wei Tao and Xi Xiao and Qianpu Sun and Yuxin Zheng and Wentao Mo and Peiqiang Wang and Nan Zhang},
      year={Wed Nov 26 2025 18:37:54 GMT+0000 (Coordinated Universal Time)},
      eprint={2511.21663},
      archivePrefix={arXiv},
      primaryClass={cs.AI},
      url={https://arxiv.org/abs/2511.21663},
}

@inproceedings{todorov2012mujoco,
  title={Mujoco: A physics engine for model-based control},
  author={Todorov, Emanuel and Erez, Tom and Tassa, Yuval},
  booktitle={2012 IEEE/RSJ international conference on intelligent robots and systems},
  pages={5026--5033},
  year={2012},
  organization={IEEE}
}

@article{liu2023libero,
  title={Libero: Benchmarking knowledge transfer for lifelong robot learning},
  author={Liu, Bo and Zhu, Yifeng and Gao, Chongkai and Feng, Yihao and Liu, Qiang and Zhu, Yuke and Stone, Peter},
  journal={Advances in Neural Information Processing Systems},
  volume={36},
  pages={44776--44791},
  year={2023}
}

@inproceedings{xiao2019meshadv,
  title={Meshadv: Adversarial meshes for visual recognition},
  author={Xiao, Chaowei and Yang, Dawei and Li, Bo and Deng, Jia and Liu, Mingyan},
  booktitle={Proceedings of the IEEE/CVF conference on computer vision and pattern recognition},
  pages={6898--6907},
  year={2019}
}

@inproceedings{huang2024towards,
  title={Towards transferable targeted 3d adversarial attack in the physical world},
  author={Huang, Yao and Dong, Yinpeng and Ruan, Shouwei and Yang, Xiao and Su, Hang and Wei, Xingxing},
  booktitle={Proceedings of the IEEE/CVF conference on computer vision and pattern recognition},
  pages={24512--24522},
  year={2024}
}

@article{laine2020modular,
  title={Modular primitives for high-performance differentiable rendering},
  author={Laine, Samuli and Hellsten, Janne and Karras, Tero and Seol, Yeongho and Lehtinen, Jaakko and Aila, Timo},
  journal={ACM Transactions on Graphics (ToG)},
  volume={39},
  number={6},
  pages={1--14},
  year={2020},
  publisher={ACM New York, NY, USA}
}

@inproceedings{suryanto2023active,
  title={Active: Towards highly transferable 3d physical camouflage for universal and robust vehicle evasion},
  author={Suryanto, Naufal and Kim, Yongsu and Larasati, Harashta Tatimma and Kang, Hyoeun and Le, Thi-Thu-Huong and Hong, Yoonyoung and Yang, Hunmin and Oh, Se-Yoon and Kim, Howon},
  booktitle={Proceedings of the IEEE/CVF international conference on computer vision},
  pages={4305--4314},
  year={2023}
}

@inproceedings{esser2021taming,
  title={Taming transformers for high-resolution image synthesis},
  author={Esser, Patrick and Rombach, Robin and Ommer, Bjorn},
  booktitle={Proceedings of the IEEE/CVF conference on computer vision and pattern recognition},
  pages={12873--12883},
  year={2021}
}

@inproceedings{athalye2018synthesizing,
  title={Synthesizing robust adversarial examples},
  author={Athalye, Anish and Engstrom, Logan and Ilyas, Andrew and Kwok, Kevin},
  booktitle={International conference on machine learning},
  pages={284--293},
  year={2018},
  organization={PMLR}
}

@article{dong2022viewfool,
  title={Viewfool: Evaluating the robustness of visual recognition to adversarial viewpoints},
  author={Dong, Yinpeng and Ruan, Shouwei and Su, Hang and Kang, Caixin and Wei, Xingxing and Zhu, Jun},
  journal={Advances in neural information processing systems},
  volume={35},
  pages={36789--36803},
  year={2022}
}

@inproceedings{ruan2023towards,
  title={Towards viewpoint-invariant visual recognition via adversarial training},
  author={Ruan, Shouwei and Dong, Yinpeng and Su, Hang and Peng, Jianteng and Chen, Ning and Wei, Xingxing},
  booktitle={Proceedings of the IEEE/CVF International Conference on Computer Vision},
  pages={4709--4719},
  year={2023}
}

@article{dziugaite2016study,
  title={A study of the effect of jpg compression on adversarial images},
  author={Dziugaite, Gintare Karolina and Ghahramani, Zoubin and Roy, Daniel M},
  journal={arXiv preprint arXiv:1608.00853},
  year={2016}
}

@inproceedings{zhang2019defending,
  title={Defending against whitebox adversarial attacks via randomized discretization},
  author={Zhang, Yuchen and Liang, Percy},
  booktitle={The 22nd International Conference on Artificial Intelligence and Statistics},
  pages={684--693},
  year={2019},
  organization={PMLR}
}

@article{xu2017feature,
  title={Feature squeezing: Detecting adversarial examples in deep neural networks},
  author={Xu, Weilin and Evans, David and Qi, Yanjun},
  journal={arXiv preprint arXiv:1704.01155},
  year={2017}
}

@inproceedings{chen2025autobreach,
  title={AutoBreach: Universal and Adaptive Jailbreaking with Efficient Wordplay-Guided Optimization via Multi-LLMs},
  author={Chen, Jiawei and Yang, Xiao and Fang, Zhengwei and Tian, Yu and Dong, Yinpeng and Yin, Zhaoxia and Su, Hang},
  booktitle={Findings of the Association for Computational Linguistics: NAACL 2025},
  pages={6777--6798},
  year={2025}
}

@article{chen2023advfas,
  title={AdvFAS: A robust face anti-spoofing framework against adversarial examples},
  author={Chen, Jiawei and Yang, Xiao and Yin, Heng and Ma, Mingzhi and Chen, Bihui and Peng, Jianteng and Guo, Yandong and Yin, Zhaoxia and Su, Hang},
  journal={Computer Vision and Image Understanding},
  volume={235},
  pages={103779},
  year={2023},
  publisher={Elsevier}
}

@article{kurakin2016adversarial,
  title={Adversarial machine learning at scale},
  author={Kurakin, Alexey and Goodfellow, Ian and Bengio, Samy},
  journal={arXiv preprint arXiv:1611.01236},
  year={2016}
}

@article{shafahi2019adversarial,
  title={Adversarial training for free!},
  author={Shafahi, Ali and Najibi, Mahyar and Ghiasi, Mohammad Amin and Xu, Zheng and Dickerson, John and Studer, Christoph and Davis, Larry S and Taylor, Gavin and Goldstein, Tom},
  journal={Advances in neural information processing systems},
  volume={32},
  year={2019}
}

@article{wang2025robosafe,
  title={RoboSafe: Safeguarding Embodied Agents via Executable Safety Logic},
  author={Wang, Le and Ying, Zonghao and Yang, Xiao and Zou, Quanchen and Yin, Zhenfei and Li, Tianlin and Yang, Jian and Yang, Yaodong and Liu, Aishan and Liu, Xianglong},
  journal={arXiv preprint arXiv:2512.21220},
  year={2025}
}

@article{fei2025libero,
  title={Libero-plus: In-depth robustness analysis of vision-language-action models},
  author={Fei, Senyu and Wang, Siyin and Shi, Junhao and Dai, Zihao and Cai, Jikun and Qian, Pengfang and Ji, Li and He, Xinzhe and Zhang, Shiduo and Fei, Zhaoye and others},
  journal={arXiv preprint arXiv:2510.13626},
  year={2025}
}

@article{intelligence2025pi_,
  title={$pi\_0.5$: a Vision-Language-Action Model with Open-World Generalization},
  author={Intelligence, Physical and Black, Kevin and Brown, Noah and Darpinian, James and Dhabalia, Karan and Driess, Danny and Esmail, Adnan and Equi, Michael and Finn, Chelsea and Fusai, Niccolo and others},
  journal={arXiv preprint arXiv:2504.16054},
  year={2025}
}

@article{yang2025mla,
  title={Mla-trust: Benchmarking trustworthiness of multimodal llm agents in gui environments},
  author={Yang, Xiao and Chen, Jiawei and Luo, Jun and Fang, Zhengwei and Dong, Yinpeng and Su, Hang and Zhu, Jun},
  journal={arXiv preprint arXiv:2506.01616},
  year={2025}
}

@inproceedings{xiang2019pointcloud,
  title={Generating 3D adversarial point clouds},
  author={Xiang, Chong and Qi, Charles R and Li, Bo},
  booktitle={Proceedings of the IEEE/CVF Conference on Computer Vision and Pattern Recognition},
  pages={9136--9144},
  year={2019}
}

@inproceedings{li2023adv3d,
  title={Adv3D: Generating 3D adversarial examples for 3D object detection in driving scenarios with NeRF},
  author={Li, Leheng and Lian, Qing and Chen, Ying-Cong},
  booktitle={Proceedings of the IEEE/CVF International Conference on Computer Vision Workshops},
  year={2023}
}

\appendix

\end{document}